\documentclass[acmtog]{acmart}
\AtBeginDocument{%
  }

\copyrightyear{2026}
\acmYear{2026}
\setcopyright{cc}
\setcctype{by}
\acmConference[SIGGRAPH Conference Papers '26]{Special Interest Group on Computer Graphics and Interactive Techniques Conference Conference Papers}{July 19--23, 2026}{Los Angeles, CA, USA}
\acmBooktitle{Special Interest Group on Computer Graphics and Interactive Techniques Conference Conference Papers (SIGGRAPH Conference Papers '26), July 19--23, 2026, Los Angeles, CA, USA}
\acmDOI{10.1145/3799902.3811200}
\acmISBN{979-8-4007-2554-8/2026/07}

\usepackage{xspace}
\usepackage{multirow}
\usepackage{colortbl}

\newcommand{\ourmodel}{\textsc{Relit-LiVE}\xspace}
\newcommand{\parahead}[1]{\noindent\textbf{#1}}

\begin{document}

\title[Relit-LiVE: Relight Video by Jointly Learning Environment Video]{\texorpdfstring{\ourmodel{}:  \underline{Reli}gh\underline{t} Video by Jointly \underline{L}earning \underline{E}nvironment \underline{Vi}deo}{Relit-LiVE: Relight Video by Jointly Learning Environment Video}}

\author{Weiqing Xiao}
\authornote{These authors contributed equally to this work.}
\authornote{Work partially done during an internship at Beijing Academy of Artificial Intelligence.}
\email{weiqing001@smail.nju.edu.cn}
\orcid{0009-0003-2548-0485}
\affiliation{%
  \institution{Nanjing University}
  \city{Suzhou}
  \country{China}
}

\author{Hong Li}
\authornotemark[1]
\authornotemark[2]
\email{link0502@buaa.edu.cn}
\orcid{0000-0002-4240-3073}
\affiliation{%
  \institution{BAAI}
  \city{Beijing}
  \country{China}
}
\affiliation{%
  \institution{BUAA}
  \city{Beijing}
  \country{China}
}

\author{Xiuyu Yang}
\authornotemark[1]
\email{gzzyyxy@gmail.com}
\orcid{0000-0002-4596-4191}
\affiliation{%
  \institution{Tsinghua University}
  \city{Beijing}
  \country{China}
}

\author{Houyuan Chen}
\email{houyuanchen111@gmail.com}
\orcid{0009-0005-4693-2326}
\affiliation{%
  \institution{The Hong Kong University of Science and Technology}
  \city{Hong Kong}
  \country{China}
}

\author{Wenyi Li}
\email{liwenyi19@mails.ucas.ac.cn}
\orcid{0000-0002-3313-8140}
\affiliation{%
  \institution{University of Chinese Academy of Sciences}
  \city{Beijing}
  \country{China}
}

\author{Tianqi Liu}
\email{tq_liu@hust.edu.cn}
\orcid{0009-0003-0718-0614}
\affiliation{%
  \institution{Huazhong University of Science and Technology}
  \city{Beijing}
  \country{China}
}

\author{Shaocong Xu}
\email{daniellesry@gmail.com}
\orcid{0000-0001-7525-0790}
\affiliation{%
  \institution{BAAI}
  \city{Beijing}
  \country{China}
}

\author{Chongjie Ye}
\email{chongjieye@link.cuhk.edu.cn}
\orcid{0000-0002-7123-0220}
\affiliation{%
  \institution{The Chinese University of Hong Kong, Shenzhen}
  \city{Shenzhen}
  \country{China}
} 

\author{Hao Zhao}
\authornote{Corresponding authors.}
\email{zhaohao@air.tsinghua.edu.cn}
\orcid{0000-0001-7903-581X}
\affiliation{%
  \institution{Tsinghua University}
  \city{Beijing}
  \country{China}
}
\affiliation{%
  \institution{BAAI}
  \city{Beijing}
  \country{China}
}

\author{Beibei Wang}
\authornotemark[3]
\email{beibei.wang@nju.edu.cn}
\orcid{0000-0001-8943-8364}
\affiliation{%
  \institution{Nanjing University}
  \city{Suzhou}
  \country{China}
}

\renewcommand{\shortauthors}{Xiao et al.}
\begin{abstract}
Recent advances have shown that large-scale video diffusion models can be repurposed as neural renderers by first decomposing videos into intrinsic scene representations and then performing forward rendering under novel illumination. While promising, this paradigm fundamentally relies on accurate intrinsic decomposition, which remains highly unreliable for real-world videos and often leads to distorted appearances, broken materials, and accumulated temporal artifacts during relighting. In this work, we present \textbf{\ourmodel{}}, a novel video relighting framework that produces physically consistent, temporally stable results without requiring prior knowledge of camera pose. Our key insight is to explicitly introduce raw reference images into the rendering process, enabling the model to recover critical scene cues that are inevitably lost or corrupted in intrinsic representations. Furthermore, we propose a novel environment video prediction formulation that simultaneously generates relit videos and per-frame environment maps aligned with each camera viewpoint in a single diffusion process. This joint prediction enforces strong geometric–illumination alignment and naturally supports dynamic lighting and camera motion, significantly improving physical consistency in video relighting while easing the requirement of known per-frame camera pose. To further enhance generalization, we introduce two complementary training strategies: (i) latent-space interpolation between relighting and rendering outputs to synthesize diverse, photorealistic multi-illumination data, and (ii) a cycle-consistent self-supervised illumination learning scheme that enforces temporal lighting coherence without additional annotations. Extensive experiments demonstrate that \ourmodel{} consistently outperforms state-of-the-art video relighting and neural rendering methods across synthetic and real-world benchmarks. Beyond relighting, our framework naturally supports a wide range of downstream applications, including scene-level rendering, material editing, object insertion, and streaming video relighting. The Project is available at \url{https://github.com/zhuxing0/Relit-LiVE}.
\end{abstract}



\begin{teaserfigure}
\centering
\includegraphics[width=\textwidth]{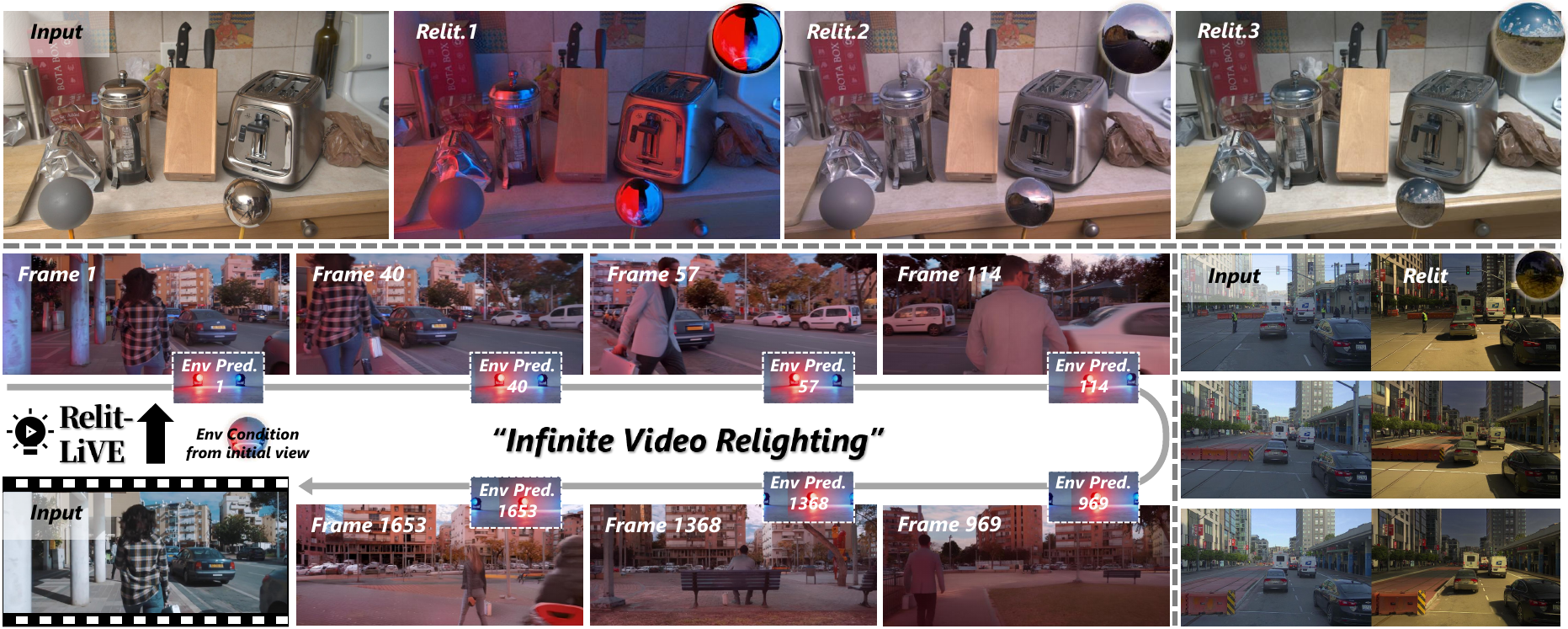}
\caption{We present \textbf{\ourmodel{}}, a novel video relighting framework that produces physically consistent and temporally stable results without needing prior knowledge of camera pose. This is achieved by jointly generating relighting videos and environment videos. Additionally, by integrating real-world lighting effects with intrinsic constraints, the relighting videos demonstrate remarkable physical plausibility, showcasing realistic reflections and shadows.
}
\label{fig:teaser}
\end{teaserfigure}

\maketitle

\section{Introduction}
\label{sec:intro}

Video relighting aims to modify a video's illumination while preserving the scene's intrinsic properties. It has various applications, including content creation, creative editing, and robust vision systems. However, it  remains a long-standing challenge to achieve physically consistent and temporally accurate lighting effects, such as realistic reflections or stable, time-coherent shadows. Addressing this requires not only accounting for different material properties but also precise, controllable modeling of lighting conditions.

Building upon powerful pre-trained diffusion models, several studies~\cite{zhou2025light,liutc} directly generate relit videos using text prompts or background images as lighting conditions. While achieving breakthroughs in visual quality, these methods typically lack precise lighting control and often retain artifacts from the original illumination. In contrast to direct generation, another line of research~\cite{liang2025diffusion,fang2025v} explores a two-stage architecture that incorporates an intermediate step of intrinsic decomposition. This approach first separates scene intrinsics from illumination, then performs relighting synthesis based on these components, using environment maps for conditioning. This explicit separation enables a clearer decoupling between scene properties and lighting, facilitating higher visual quality and more precise control. However, this paradigm is heavily dependent on the fidelity of the intermediate intrinsic representation. In challenging scenarios, such as transparent objects with complex light transport or subsurface scattering, neural intrinsic rendering might yield flawed or implausible outputs. A recent work by He et al.~\shortcite{heunirelight} unifies albedo estimation with direct relighting, synthesizing scene albedo and relighting video in parallel to effectively decouple and reshape scene illumination. However, constrained by the inherent challenges of training parallel inference paradigms, their approach struggles to extend to more intrinsic properties, limiting its capabilities. Furthermore, these methods require precise prior knowledge of the video camera's pose to position the environment map in the viewport, which constrains their flexibility. 

In this paper, we propose \ourmodel{}, a novel video relighting framework that produces physically consistent, temporally stable results without requiring prior knowledge of camera pose. To this end, we address two core challenges: (1) preserving scene content integrity under complex light transport, and (2) flexibly injecting novel lighting conditions without known camera pose. We present two key insights to address these challenges. First, while decomposed intrinsic attributes often struggle to capture complex global illumination effects, these effects are directly observable in the original RGB video sequence. Therefore, we propose an RGB-intrinsic fusion renderer that utilizes the input RGB frames—also known as raw reference images—to guide and correct the rendering process, providing both visual and semantic-level cues. This design fuses the RGB space with the intrinsic space, enabling the model to incorporate real-world lighting effects alongside estimated physical constraints, resulting in realistic relighting results. Second, to facilitate arbitrary relighting without requiring per-frame camera poses, we reformulate relit video learning as the simultaneous learning of a per-frame warping of the environment map in combination with relit video synthesis. This approach allows our model to generate both relit videos and per-frame warped environment maps (referred to as environment video) during a single inference pass. By inferring the lighting transformation implicitly, our approach eliminates the need for explicit pose estimation, enhancing practical flexibility.

Furthermore, we improve the robustness of our model to handle complex scenarios by enhancing the training data in two ways. First, we perform latent-space interpolation between relighting and rendering outputs using the initially trained model. This allows us to synthesize diverse, photorealistic multi-illumination data. Second, we employ a cycle-consistent self-supervised illumination learning scheme that ensures temporal lighting coherence without the need for additional annotations.

Extensive experiments demonstrate that \ourmodel{} outperforms existing state-of-the-art methods, achieving realistic material reflection effects and effectively modeling viewpoint changes in videos. This enables us to perform physically plausible and spatio-temporally accurate relighting of videos without requiring camera pose priors. \ourmodel{} also offers flexibility for task extension, enabling scene-level rendering, editing, and streaming video relighting through modifying generation conditions and intermediate outputs.
In summary, our contributions are as follows:
\begin{itemize}
  \item a novel video relighting framework, \ourmodel{}, that produces physically consistent, temporally stable results without requiring prior knowledge of camera pose,
  \item an RGB-intrinsic fusion renderer, that effectively integrates real-world lighting effects from the RGB space with physical constraints from the intrinsic space, enabling the generation of physically consistent video lighting effects, and
  \item jointly generation of relit video and environment video, enabling geometry-illumination aligned video relighting without requiring per-frame camera poses.
\end{itemize}

\section{Related work}
\label{sec:related}

\begin{figure*}[t]
  \centering
  \includegraphics[width=\textwidth]{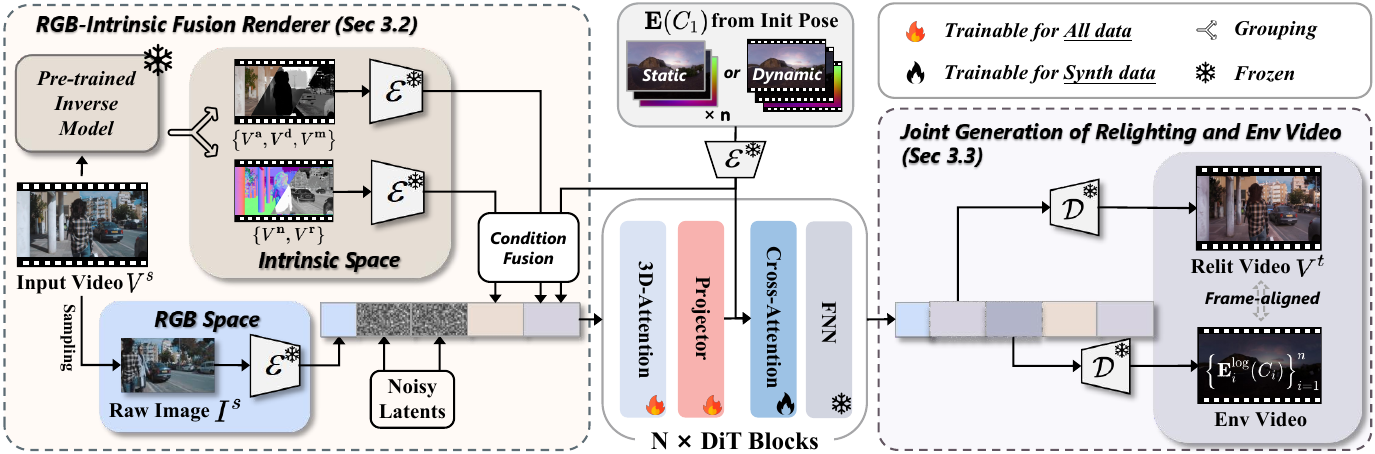}
  \caption{\textbf{Overview of our \ourmodel{}.} Given an input video and environment maps of the initial viewpoint, our method jointly predicts relit videos and frame-specific environment maps (i.e., environment video). The input video is converted into intrinsic properties by a pre-trained inverse rendering model, then mapped into latent space alongside environment maps and randomly sampled reference images. Subsequently, latents undergo partial grouping fusion and frame-wise concatenation, followed by denoising through the DiT video model to generate realistic relighting video.
  }
   \label{fig:main}
\end{figure*}

\subsection{Direct video relighting} 

Direct video relighting aims to adjust the lighting conditions of a video while preserving the scene content through an end-to-end approach.
Driven by breakthroughs in controllable video diffusion technology~\cite{wan2025,yangcogvideox}, this paradigm has achieved rapid development.
Overall, the research focus of this paradigm is shifting from the mere pursuit of temporal consistency toward precise lighting control and physical realism.

Some early studies~\cite{fang2025relightvid,fang2025relightvid,liutc} have focused on achieving temporally consistent relighting, typically using text prompts or reference backgrounds as rough lighting conditions.
For instance, methods such as Light-A-Video~\cite{zhou2025light} and TC-Light~\cite{liutc} extend the effects of the image re-illumination technique IC-Light~\cite{zhang2025scaling} smoothly across entire videos through carefully designed temporal consistency enhancement schemes.
Recent research~\cite{ren2025mv,liulight,magar2025lightlab} has increasingly focused on precise control and physical realism in lighting, with representative methods including RelightMaster~\cite{bian2025relightmaster}, UniLumos~\cite{liuunilumos}, and UniRelight~\cite{heunirelight}.
RelightMaster~\cite{bian2025relightmaster} and UniLumos~\cite{liuunilumos} respectively propose multi-plane light images and structured text prompts to achieve fine-grained control over lighting parameters.
Additionally, UniLumos incorporates depth and normal geometric feedback supervision to ensure shadow plausibility. UniRelight~\cite{heunirelight} jointly learns to directly generate relit videos and albedo estimation. By implicitly decoupling ambient lighting, it enhances lighting effects in complex scenes.

However, this parallel inference pattern presents inherent training challenges: model capacity often limits the scope of tasks it can handle. This constrains the upper bound of the joint estimation paradigm, making it difficult to account for comprehensive intrinsic properties.
In contrast, our decoupled approach ensures both the comprehensiveness and expandability of intrinsic content.
This also grants our method greater architectural flexibility, supporting not only video relighting but also tasks like neural rendering.

\subsection{Intrinsic-aware diffusion model} 

Inspired by Physically-Based Rendering (PBR) pipelines~\cite{rendering2015physically}, some research~\cite{beisswenger2025framediffuser,kocsisintrinsix,ye2024stablenormal} has begun exploring the intrinsic decomposition~\cite{careaga2023intrinsic,bonneel2017intrinsic,shu2018deforming} and synthesis of images and videos through diffusion models~\cite{chen2025physgen3d}. Compared to end-to-end generation, this paradigm offers high flexibility.
By adjusting its intrinsic components, it can perform a variety of functions, including light modification and material editing.

Some approaches~\cite{kocsisintrinsix,chen2025invrgb+,helotus,careaga2025physically} focus on intrinsic decomposition tasks, with representative methods including IntrinsiX~\cite{kocsisintrinsix}, NormalCrafter~\cite{bin2025normalcrafter}, and GeometryCrafter~\cite{xu2025geometrycrafter}. These methods are based on fine-tuning pre-trained diffusion models. Leveraging the strong generative prior of diffusion models, they achieve precise decomposition of specific intrinsic properties through conditional generation.
Other studies~\cite{liang2025diffusion,fang2025v,chen2025uni,xi2025ctrlvdiff} simultaneously focus on both intrinsic decomposition and synthesis tasks to achieve a closed-loop ``decomposition-synthesis'' capability.
For instance, RGBX~\cite{zeng2024rgb} employs image diffusion models to enable bidirectional functionality: estimating G-buffers from images and rendering images based on G-buffers.
Recent work such as the Diffusion Renderer~\cite{liang2025diffusion} and V-RGBX~\cite{fang2025v} extends this closed-loop architecture from images to the video domain. However, constrained by the inherent challenges of decomposing intrinsic properties in the real world, this ``decomposition-synthesis'' architecture is often limited to specific domains and prone to cumulative error issues. Additionally, during the compositing stage, such methods typically require precise lighting information, such as irradiance maps or environment maps for all frames. This limits the practicality of its relighting function. In our paper, we propose a novel video relighting framework with two key designs to address the two challenges outlined above.

\section{Our method}
\label{sec:method}

This paper targets the problem of video relighting, aiming to generate physically consistent and temporally stable results without relying on prior camera pose estimation. In this section, we first formalize the problem and then introduce our proposed framework, \ourmodel{}, as shown in Figure~\ref{fig:main}.

\subsection{Problem statement}

For the task of video relighting, we are given a source video sequence \( V^s = \{ I^s_i \}_{i=1}^n \) $\in\mathbb{R}^{n\times h\times w \times 3}$ and a target lighting sequence \( \mathbf{E}^t = \{ \mathbf{E}_i \}_{i=1}^n \) $\in \mathbb{R}^{n\times h \times w \times 3}$ (which may be static or dynamic). The objective is to synthesize a target video \( V^t = \{ I^t_i \}_{i=1}^n \) that faithfully exhibits the original scene content from \( V^s \) under the novel illumination \( \mathbf{E}^t \), effectively replacing the source lighting. This process can be formulated as:
\begin{equation}
V^t = \mathcal{F_\theta} (V^s, \mathbf{E}^t),
\label{eqn:task}
\end{equation}
where \( \mathcal{F_\theta} \) is a relighting network parameterized by \( \theta \). In the case of static target lighting, the sequence \( \mathbf{E}^t \) reduces to a constant environment map applied to every frame.

\subsection{RGB-Intrinsic fusion renderer}
\label{sec:model_design}

Learning the video relighting task directly is challenging because it is inherently difficult to disentangle the intrinsic scene properties from the original lighting conditions. Hence, a common paradigm in video relighting involves first performing an intrinsic decomposition of the source video to separate material properties from illumination, followed by re-rendering the extracted materials under the target lighting. In this view, the renderer serves as a relighting pathway. This paradigm improves physical plausibility, but its performance is critically limited by the accuracy and robustness of the decomposition stage. This limitation becomes particularly apparent in scenes with complex lighting effects, leading to visual artifacts. Thus, the reliance on imperfect intrinsic decomposition remains a core challenge in achieving high-fidelity video relighting. To resolve this issue, we find that these lighting effects are directly observable in the original RGB video. The raw images provide visual and even semantic-level cues for video rendering tasks in RGB space, while intrinsic properties in G-buffer impose direct physical constraints on relighting results. Therefore, we propose an RGB-Intrinsic fusion renderer, which utilizes this observable RGB information to guide the rendering process, thus bypassing the limitations posed by imperfect intrinsic decomposition.

Given a source video $V^s$, we utilize the inverse renderer from Diffusion Renderer~\cite{liang2025diffusion} to predict its G-buffers, which include a common set of intrinsic properties: base color $V^\mathbf{a}$, surface normal $V^\mathbf{n}$, relative depth $V^\mathbf{d}$, roughness $V^\mathbf{r}$, and metallic $V^\mathbf{m}$). We then employ a pretrained VAE encoder $\mathcal{E}$ to encode each G-buffer into te latent space, resulting in the corresponding latents $\left \{ \mathbf{z} ^\mathbf{a} ,\mathbf{z}^\mathbf{n} ,\mathbf{z}^\mathbf{d} ,\mathbf{z}^\mathbf{r} ,\mathbf{z}^\mathbf{m} \right \}$, where $\mathbf{z} ^\mathbf{\ast }\in \mathbb{R}^{N\times H \times W \times C}$.

Previous works~\cite{liang2025diffusion,zeng2024rgb,fang2025v} have directly concatenated these intrinsic latents either along the frame or channel dimension. However, we have observed that concatenating along the frame dimension increases computational overhead, while concatenating along the channel dimension slows down model convergence. To address these issues, we propose to sum the latents partially before concatenating them along the frame dimension. From a pilot study, we identified a key point: separating intrinsic properties that exhibit similar numerical characteristics or strong correlations—such as metallic and roughness, or depth and normal—facilitates precise control over the generated results. The former two are typically represented by grayscale values and demonstrate pronounced regional equivalence, meaning regions with the same material tend to maintain nearly constant values; the latter two exhibit significant numerical correlation. Therefore, we specifically separate these modalities during G-buffer grouping. Specifically, we compute two new sets of latents: $\mathbf{z} ^{\left \{\mathbf{a},\mathbf{d},\mathbf{m}\right \}} = \mathbf{z} ^\mathbf{a}+\mathbf{z}^\mathbf{d}+\mathbf{z}^\mathbf{m}$ and $\mathbf{z} ^{\left \{\mathbf{n},\mathbf{r}\right \}} = \mathbf{z} ^\mathbf{n}+\mathbf{z}^\mathbf{r}$. These two new latents serve as intrinsic conditions. 

Then, we randomly sample a raw image $I^s$ from the input video and use the VAE encoder $\mathcal{E}$ to encode this image, generating the latent $\mathbf{z} ^\mathbf{I} \in \mathbb{R}^{1\times H \times W \times C}$. This latent representation is concatenated with intrinsic conditions along the frame dimension, effectively guiding the generation process together. This random sampling strategy breaks fixed correspondences between the raw image and generated results, thereby suppressing pixel-level propagation of source lighting. It is worth noting that, since the inference process of diffusion models typically involves multiple denoising steps, we can actually sample different frames during each denoising step to preserve as much detail as possible.

\subsection{Joint generation of relighting and environment video}

With the encoded features and environment maps, we could render them using a DiT video model to generate the relit video. Since \(\mathcal{F_\theta}\) operates in 2D image space, the environment maps \( \{ \mathbf{E}_i \}_{i=1}^n \) must be appropriately aligned with the camera's viewing direction. 
Here, we set $\{ \mathbf{E}_i \}_{i=1}^n =\left \{\mathbf{E}_i(C_i) \right \} ^n_{i=1}$ to highlight this operation, where $C_i$ represents the i-th camera viewpoint. While the source video inherently defines the camera poses, these poses are often unknown or inaccurately estimated in practice. Existing methods often assume known camera poses, allowing for direct warping of the environment map into camera space. However, this assumption limits their real-world applicability. To address this issue, we propose learning warped environment maps (referred to herein as environment videos) along with the relit video. This way, the DiT model can be forced to learn render the scene with the warped environment maps. By implicitly inferring lighting transformations, we eliminate the need for explicit pose estimation, enhancing practical usability while ensuring spatio-temporal lighting accuracy.

We start by reformulating our relight task into the joint generation of the relit video and the warped environment video.
\begin{equation}
\begin{split}
V^t, \left\{ \mathbf{E}_i(C_i) \right\}_{i=1}^n  &= \mathcal{F}_\theta \left( V^s, \left\{ \mathbf{E}_i(C_1) \right\}_{i=1}^n \right) \\
&= \mathcal{F}_\theta \left( I^s, V^\mathbf{a}, V^\mathbf{n}, V^\mathbf{d}, V^\mathbf{r}, V^\mathbf{m}, \left\{ \mathbf{E}_i(C_1) \right\}_{i=1}^n \right).
\end{split}
\label{eqn:formulation_combined}
\end{equation}
In the above equation, we also incorporate intrinsic properties along with the raw reference image introduced in the previous section. Next, we describe our lighting conditions, followed by the joint generation.

We use HDR environment maps $\mathbf{E}(C_1)$ under the initial viewpoint $C_1$ to represent lighting condition (which may be static or dynamic). Inspired from prior works~\cite{liang2025diffusion}, we construct three complementary representations for HDR environment maps: 1) LDR images $\mathbf{E}^\mathrm{ldr}(C_1)$ obtained via Reinhard tonemapping; 2) normalized log-intensity images $\mathbf{E}^\mathrm{log}(C_1) = \mathrm{log}(1+\mathbf{E}(C_1))/\mathrm{log}(1+M)$, where $M=\mathrm{60000}$; 3) directional encoding images $\mathbf{E}^\mathrm{dir}$, where each pixel represents the direction of the corresponding ray in the camera coordinate system (note that the pixel direction here is opposite to that in standard panoramas). 
We use the VAE encoder $\mathcal{E}$ to encode these three representations into the latent space separately and concatenate them along the channel dimension to obtain $\mathbf{h_E} = \left \{ \mathcal{E}(\mathbf{E}^\mathrm{ldr}(C_1)), \mathcal{E}(\mathbf{E}^\mathrm{log}(C_1)), \mathcal{E}(\mathbf{E}^\mathrm{dir})\right \}\in \mathbb{R}^{N\times H \times W \times 3C}$. Then, we process the $\mathbf{h_E}$ using a convolutional layer with a stride of 1 to obtain $\mathbf{c_E}\in \mathbb{R}^{N\times H \times W \times C}$, which is concatenated with other conditional latents. Additionally, we repeat this process at an input resolution of $512\times 256$, feeding the result $\mathbf{c}_\mathbf{E}^{\mathrm{cross}}$ separately into the cross-attention module as enhanced lighting control. 

Then, our simultaneously generates relit video $V^t$ and corresponding environment video (in the form of normalized log intensity maps $\left \{  \mathbf{E}^\mathrm{log}_i(C_i) \right \} ^n_{i=1}$, as they can be inverse-transformed back to HDR and LDR maps) using multiple DiT blocks.
During training, we encode both into the latent space using the VAE encoder $\mathcal{E}$, yielding $\mathbf{z}^\mathbf{t}$ and $\mathbf{z}^\mathbf{E_{log}}$. Subsequently, noise is independently introduced to generate $\mathbf{z}^\mathbf{t}_\tau$ and $\mathbf{z}^\mathbf{E_{log}}_\tau $. Next, we concatenate these noise-added target latents with the reference latent $\mathbf{z} ^\mathbf{I}$, intrinsic latents $\left \{ \mathbf{z} ^{\left \{\mathbf{a},\mathbf{d},\mathbf{m}\right \}},\mathbf{z} ^{\left \{\mathbf{n},\mathbf{r}\right \}} \right \} $, and lighting conditions $\left \{ \mathbf{c_E},\mathbf{c}_\mathbf{E}^{\mathrm{cross}} \right \} $ at the frame level, and feed them into DiT blocks to learn denoising:
\begin{equation}
\hat{\mathbf{z}}^\mathbf{t}(\theta ),\hat{\mathbf{z}}^\mathbf{E_{log}}(\theta ) = \mathbf{f}_\theta  ([\mathbf{z}^\mathbf{I},\mathbf{z}_{\tau }^\mathbf{t},\mathbf{z}_{\tau }^\mathbf{E_{log}},\mathbf{z} ^{\left \{\mathbf{a},\mathbf{d},\mathbf{m}\right \}},\mathbf{z} ^{\left \{\mathbf{n},\mathbf{r}\right \}}+\mathbf{c_E}];\mathbf{c}_\mathbf{E}^{\mathrm{cross}},\tau ),
\end{equation}
where [·] denotes concatenation in the temporal dimension, and $\mathbf{f}_\theta$ is the denoising function of DiT blocks. 

\subsection{Training strategies}
\label{sec:traing_strategy}

The training of our method can be divided into three stages. In the first stage, we train the model using standard supervised learning (see supplemental material for data generation strategy and training details) to acquire basic relighting capabilities. In the second and third stages, we introduce two strategies to enhance generalization:

\begin{figure}[t]
  \centering
  \includegraphics[width=0.4\textwidth]{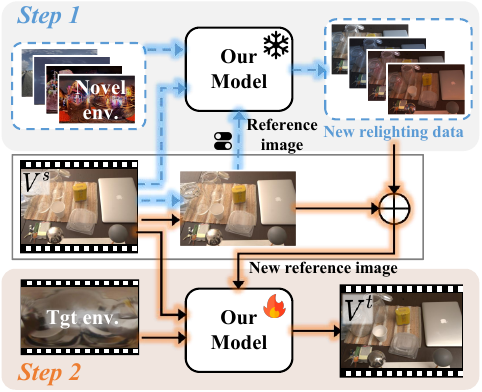}
  \caption{\textbf{Overview of intrinsic perception enhancement.} Step 1: Generate multi-illumination data. Step 2: Use these multi-illumination data as the raw reference images for training.}

\label{fig:main_train_2}
\end{figure}

\paragraph{Intrinsic perception enhancement.}
As shown in Figure~\ref{fig:main_train_2}, we randomly select environment maps and generate two relighting results by controlling whether latent $\mathbf{z}^\mathbf{I}$ is set to 0. Ideally, these two inference modes should produce identical outcomes for the same scene. But that is not the case. Overall, we found that using $\mathbf{z}^\mathbf{I}$ yields more realistic appearances but occasionally retains source lighting, whereas the variant with $\mathbf{z}^\mathbf{I}$ set to 0 avoids residual lighting but suffers from detail distortion due to cumulative errors in the inverse rendering process. Therefore, We interpolate these two results in the latent space and decode the interpolated outputs, yielding a large amount of pseudo-realistic relit data. This process can be formulated as:
\begin{equation}
\label{interpolation}
\mathbf{z}_{\mathrm{new}} = \frac{\mathbf{z}_{\mathrm{w/}}}{1+w} +\frac{\mathbf{z}_{\mathrm{w/o}}*w}{1+w}, 
\end{equation}
where $w$ is the interpolation weight, $\mathbf{z}_{\mathrm{w/}}$ denotes the latents corresponding to results with $\mathbf{z}^\mathbf{I}$, and $\mathbf{z}_{\mathrm{w/o}}$ denotes those with $\mathbf{z}^\mathbf{I}$ set to 0. Subsequently, we decode the interpolated latent $\mathbf{z}_{\mathrm{new}}$ using the VAE decoder $\mathcal{D}$ to obtain new data that trade off realism and lighting plausibility. Furthermore, we treat these data as new raw reference images to enable training under diverse lighting conditions on real-world scenes. This strategy allows our method to access a wide variety of novel lighting conditions on real-world scenes during training, thereby significantly enhancing its perception of image intrinsic properties.

\begin{figure}[t]
  \centering
  \includegraphics[width=0.45\textwidth]{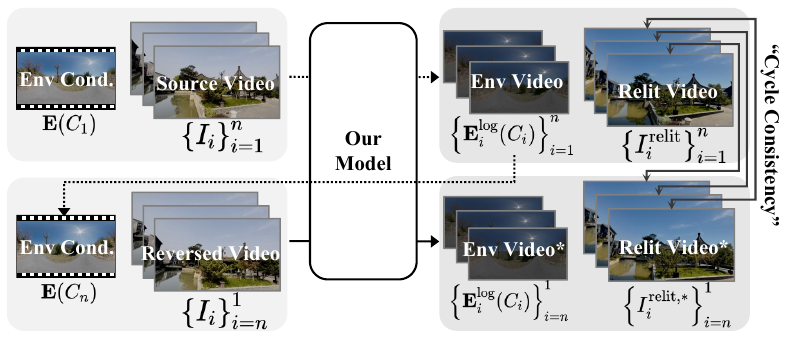}
  \caption{\textbf{Overview of self-supervised learning based on illumination consistency.} The symbol (*) denotes inference results for reverse-order video.} Dotted line operations do not compute gradients. We relit a video under random environment map and then relit the video in reverse order based on the final frame of generated environment video. These two relit results form a self-supervised training pair.
   \label{fig:main_train_3}
\end{figure}

\paragraph{Self-supervised learning based on illumination consistency.}
In the final training stage, we introduce a self-supervised illumination consistency (SIC) strategy to enhance the model's generalization across diverse scenes and lighting conditions, as illustrated in Figure~\ref{fig:main_train_3}. Specifically, we perform inference on all data under random environment maps to obtain relit video $V^{\mathrm{relit} }=\left \{  I^\mathrm{relit}_i \right \} ^n_{i=1}$ and their corresponding environment light $ \left \{ \mathbf{E}^\mathrm{log}_i(C_i) \right \} ^n_{i=1}$. We then reverse the frame sequence of the original video and infer the new relighting result $V^{\mathrm{relit,\ast} }=\left \{  I^\mathrm{relit,\ast }_i \right \} ^1_{i=n}$ based on the environmental light $\mathbf{E}^\mathrm{log}_n(C_n)$. Self-supervised training pairs are constructed through frame-to-frame correspondence. This self-supervised process operates on image data under the ``lighting rotation with fixed camera'' pattern. The SIC strategy exposes our method to diverse lighting and scene combinations, significantly improving its generalization performance. It also promotes frame-by-frame alignment between predicted lighting and relit results, enhancing its generalization and sensitivity to varying lighting conditions.
\section{Results}
\label{sec:experiments}

We compare \ourmodel{} with various advanced video relighting methods, including UniRelight~\cite{heunirelight}, Diffusion Renderer (cosmos)~\cite{liang2025diffusion}, Light-A-Video~\cite{zhou2025light}, and others.
Evaluation data spans multiple domains—synthetic, human \cite{pexels2025}, embodied~\cite{walke2023bridgedata}, and autonomous driving~\cite{xiao2021pandaset}—encompassing over 1,400 dynamic videos.
Metrics encompass visual fidelity (PSNR, SSIM, and LPIPS), temporal consistency (RAFT score), and specially designed material fidelity (DINOv3 score), supplemented by user study.
More experimental settings and results are detailed in the supplemental material.

\subsection{Evaluation of video relighting}
\label{sec:video_relight}

\begin{table*}[t]
\caption{\textbf{Quantitative comparison of relighting on the synthetic dataset and MIT multi-illumination dataset.} (*) indicates that metrics are sourced from the reported results of UniRelight~\cite{heunirelight}. Our approach surpasses the baselines across all test metrics.}
\label{tab:eval_relight_1}
\resizebox{0.95\linewidth}{!}{%
\begin{tabular}{l|ccc|ccc|ccc}
\toprule[1pt]
\multirow{2}{*}{Methods} & \multicolumn{3}{c|}{\textit{Synthetic Image}}     & \multicolumn{3}{c|}{\textit{Synthetic Video}}    & \multicolumn{3}{c}{\textit{MIT multi-illumination}}            \\ \cmidrule{2-10} 
                         & PSNR ($\uparrow $) & SSIM ($\uparrow $) & LPIPS ($\downarrow$) & PSNR ($\uparrow $) & SSIM ($\uparrow $) & LPIPS ($\downarrow$) & PSNR ($\uparrow $) & SSIM ($\uparrow $) & LPIPS ($\downarrow$) \\ \midrule[0.5pt]
NeuralGaffer~\cite{jin2024neural}             &  12.84           &   0.435           &  0.463            & -                  & -                  & -                    & 17.87*              & 0.683*              & 0.241*                \\
Diffusion Renderer~\cite{liang2025diffusion}        & 17.09            & 0.679             & 0.264             & 16.45            & 0.665            & 0.278              & 17.29*              & 0.622*              & 0.355*                \\
UniRelight~\cite{heunirelight}               & -                  & -                  & -                    & -                  & -                  & -                    & 20.76*              & 0.749*              & 0.251*                \\
Ours                     & \textbf{24.85}            & \textbf{0.792}            & \textbf{0.175}              & \textbf{25.39}            & \textbf{0.807}            & \textbf{0.205}              & \textbf{21.86}     & \textbf{0.849}     & \textbf{0.132}       \\ \bottomrule[1pt]
\end{tabular}
}
\end{table*}

\begin{figure}[t]
  \centering
  \includegraphics[width=0.48\textwidth]{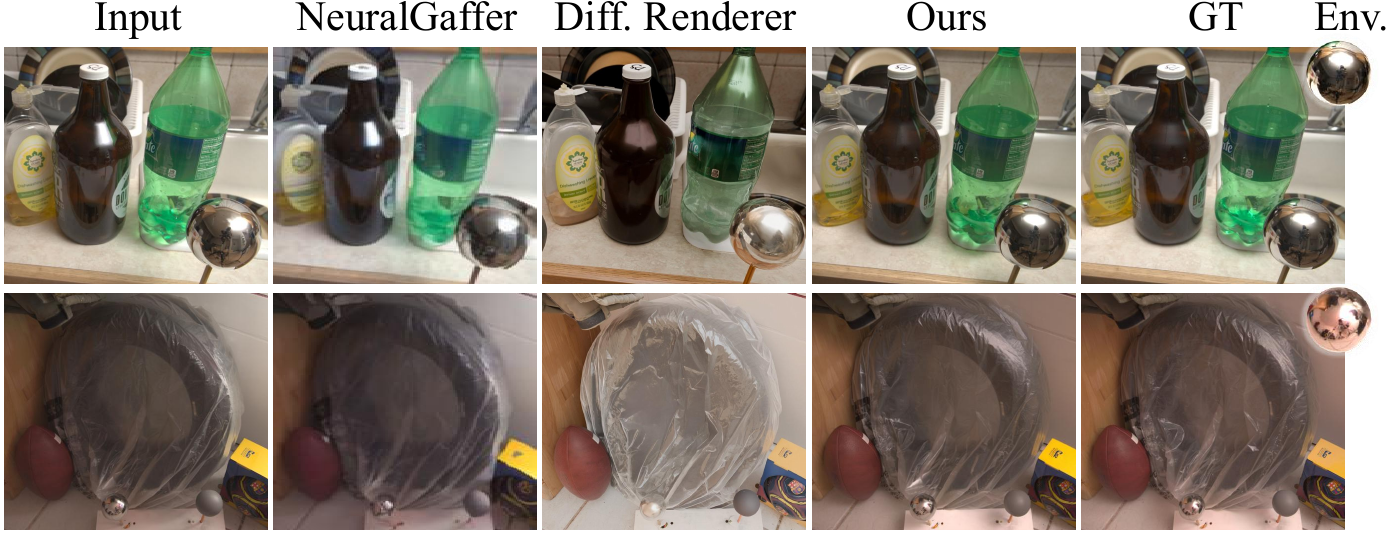}
  \caption{\textbf{Qualitative comparison of image relighting on the MIT multi-illumination dataset.} Our method excels in handling complex materials, generating high-quality reflection and transmission effects that significantly outperform baselines.}
   \label{fig:eval_relight_1}
\end{figure}

\begin{figure*}[t]
  \centering
  \includegraphics[width=0.95\textwidth]{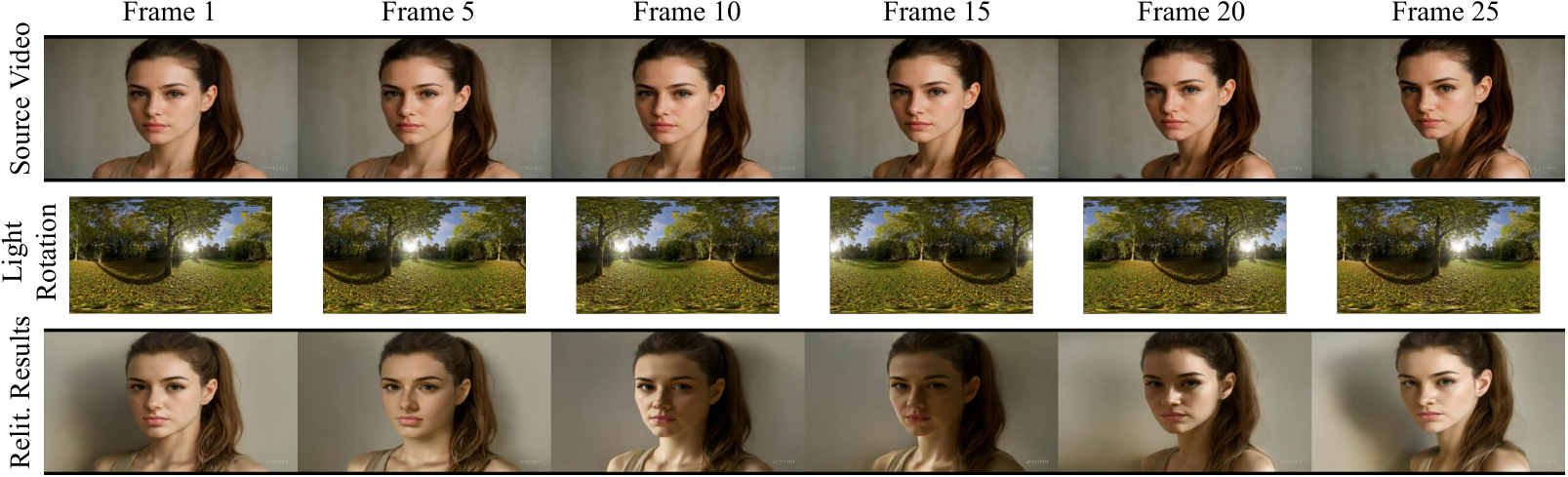}
  \caption{\textbf{Results under dynamic lighting in a dynamic scene.} Our method remains stable under simultaneous changes in scene content and illumination.}
  \label{fig:dynamic_lighting}
\end{figure*}

We compare \ourmodel{} with existing advanced methods across different datasets, with quantitative results presented in Table~\ref{tab:eval_relight_1}.
Figure~\ref{fig:eval_relight_1} and supplemental material present corresponding visualizations. 
Among them, NeuralGaffer fails on scene-level tests, struggling to remove lighting details such as shadows and highlights from the original scene.
Diffusion Renderer exhibits distortion on materials, which is particularly noticeable on transparent objects.
In contrast, our method outperforms others across all metrics while demonstrating excellent material consistency and physically accurate reflections and refractions. We also present the video relighting results of our method under dynamic lighting in Figure~\ref{fig:dynamic_lighting} and the supplemental material. 

\begin{table*}[]
\caption{\textbf{Quantitative comparison of video relighting on in-the-wild data.} Our approach significantly outperforms the baseline in terms of material consistency. User study metrics include VR (Visual Realism), PC (Physical Consistency), and LA (Lighting Alignment), reported as the percentage of participants who prefer our method. Details of each metric are provided in the supplemental material. } 
\label{tab:eval_relight_2}
\resizebox{0.95\linewidth}{!}{%
\begin{tabular}{l|c|c|cc|ccc}
\toprule[1pt]
\multirow{2}{*}{Methods} & \multirow{2}{*}{Light Condi} & Temporal Consistency & \multicolumn{2}{c|}{Material Consistency} & \multicolumn{3}{c}{User Study (\%)} \\ \cmidrule{3-8} 
& & Motion Preservation ($\downarrow$) & CLIP-MC ($\uparrow $) & DINO-MC ($\uparrow $) & VR    & PC    & LA    \\ \midrule[0.5pt]

Light-A-Video~\cite{zhou2025light} & Text & 0.4557 & 0.9150 & 0.8919 & 65.5 & 75.8 & 54.8 \\
TC-Light~\cite{liutc} & Text & 0.2405 & 0.8977 & 0.8825 & 84.5 & 84.8 & 77.4 \\

Diffusion Renderer~\cite{liang2025diffusion} & HDR & 0.3094 & 0.9105 & 0.8754 & 86.2 & 81.3 & 63.3 \\
Ours & HDR & \textbf{0.1692} & \textbf{0.9246} & \textbf{0.9091} & / &  / &  / \\ \bottomrule[1pt]
\end{tabular}
}
\end{table*}

\begin{figure*}[t]
  \centering
  \includegraphics[width=\textwidth]{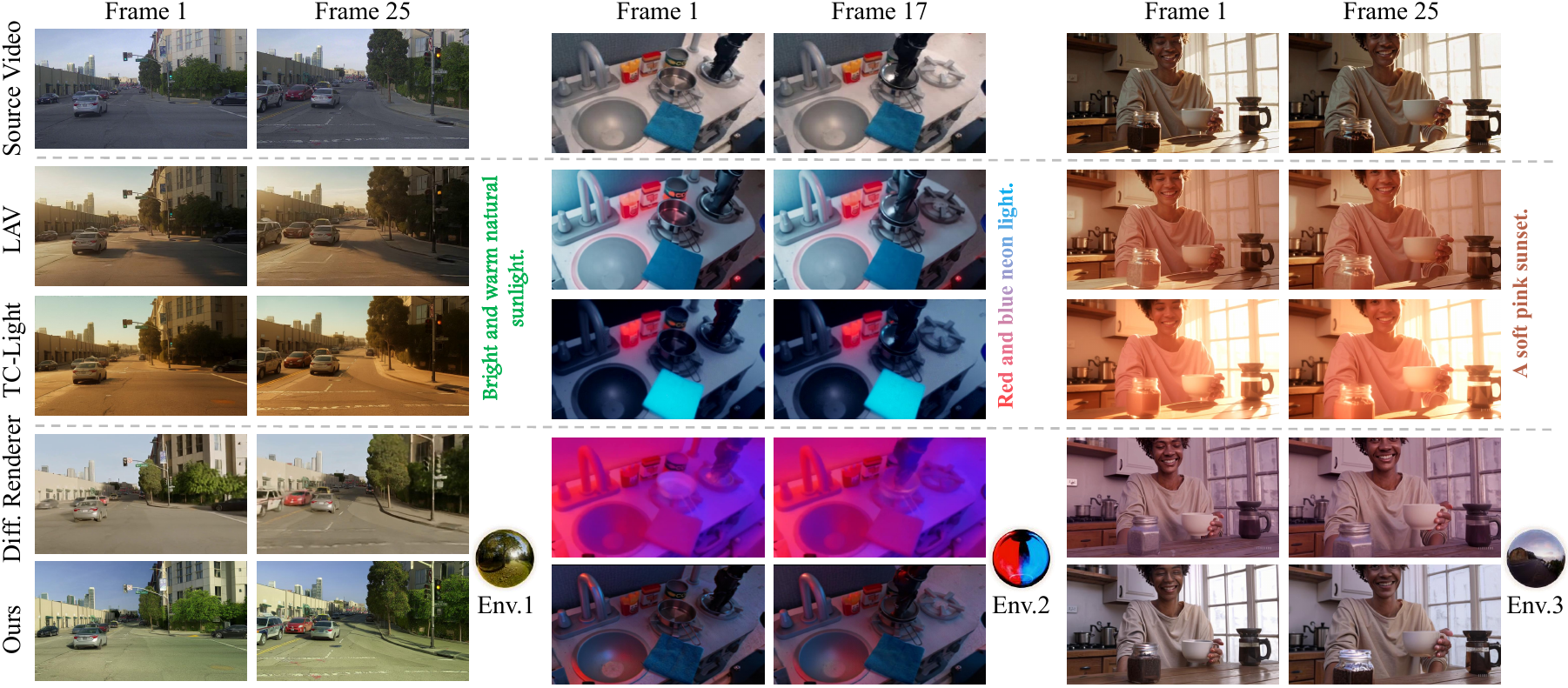}
  \caption{\textbf{Qualitative comparison of video relighting on in-the-wild data.} We simultaneously evaluate advanced environment map-based methods and text prompt-based methods, aligning their lighting styles. Our approach outperforms baselines in both relighting quality and physical consistency.}
  \label{fig:eval_relight_2}
\end{figure*}

\begin{figure*}[t]
  \centering
  \includegraphics[width=\textwidth]{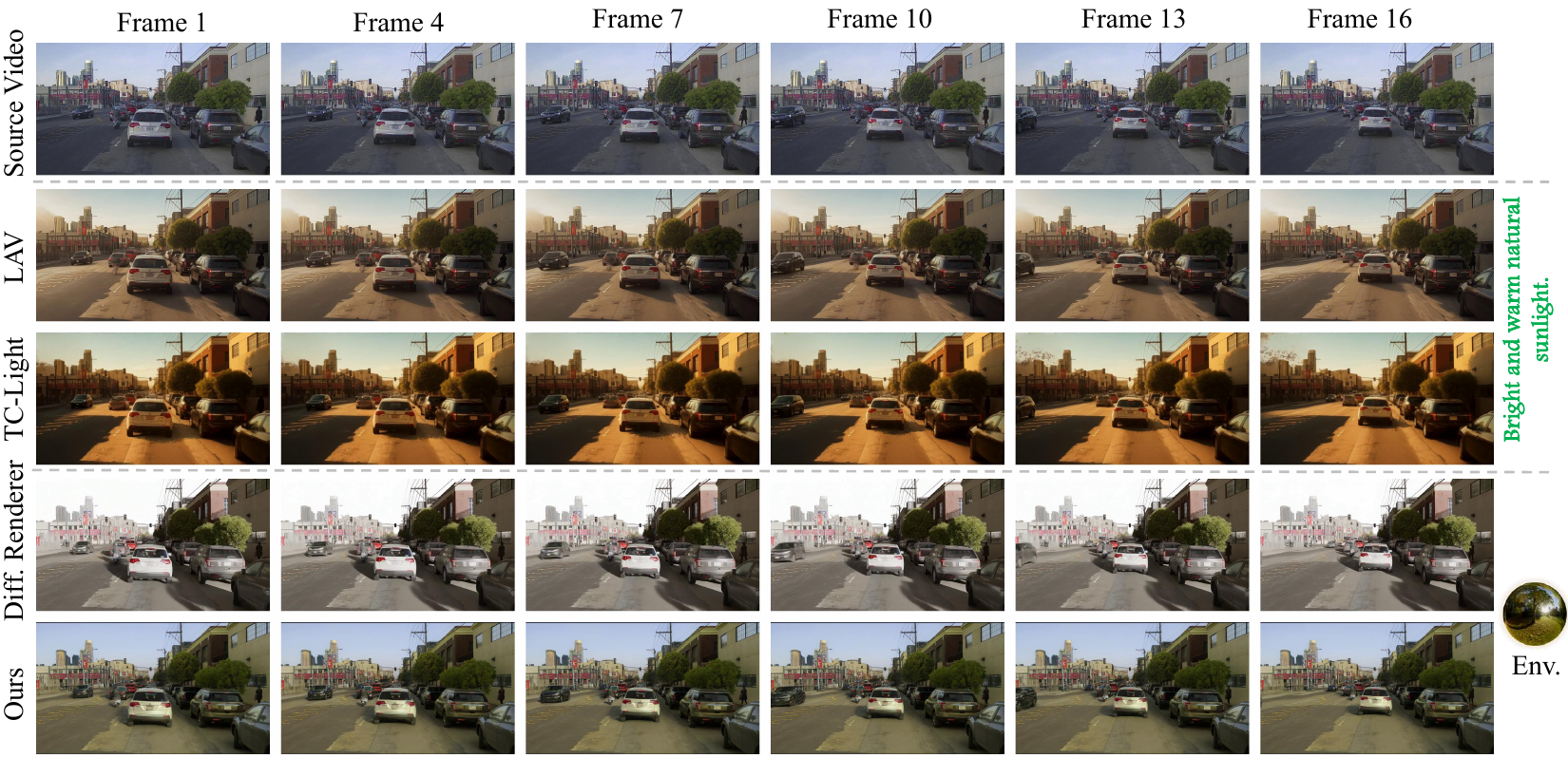}
  \caption{\textbf{Qualitative comparison of video relighting}. Our method achieves superior relighting quality, temporal consistency, and photorealistic generation results compared to baseline methods.}
  \label{fig:eval_relight_3}
\end{figure*}

Additionally, we compare \ourmodel{} with advanced text prompt-based methods across multiple domains in Table~\ref{tab:eval_relight_2}, Figure~\ref{fig:eval_relight_2} and Figure~\ref{fig:eval_relight_3}.
As shown in the middle example of Figure~\ref{fig:eval_relight_2}, due to the lack of physical constraints, text-prompt-based methods may produce unreasonable luminous effects under certain special lighting conditions, such as neon lighting.
Consequently, these methods exhibit poor material consistency, particularly evident in the DINO-MC metric.
Additionally, such methods struggle to decouple the original lighting, such as the shadows and highlights in the third example shown in the figure.
The Diffusion Renderer again exhibits material distortion due to cumulative errors in the two-stage process.
In contrast, our method demonstrates comprehensive performance, achieving both material consistency and more details in lighting and shadows.

As demonstrated in Figure~\ref{fig:teaser}, our method supports streaming video relighting.
Specifically, we can segment the long video into multiple clips. Given the lighting conditions, we perform relighting starting from the first clip. Then, based on the generated environment video, we provide the lighting conditions for the first frame's viewpoint of the next clip, performing relighting clip by clip. This allows us to naturally achieve relighting for long videos.

\begin{figure*}[t]
  \centering
  \includegraphics[width=\textwidth,height=0.35\textheight,keepaspectratio]{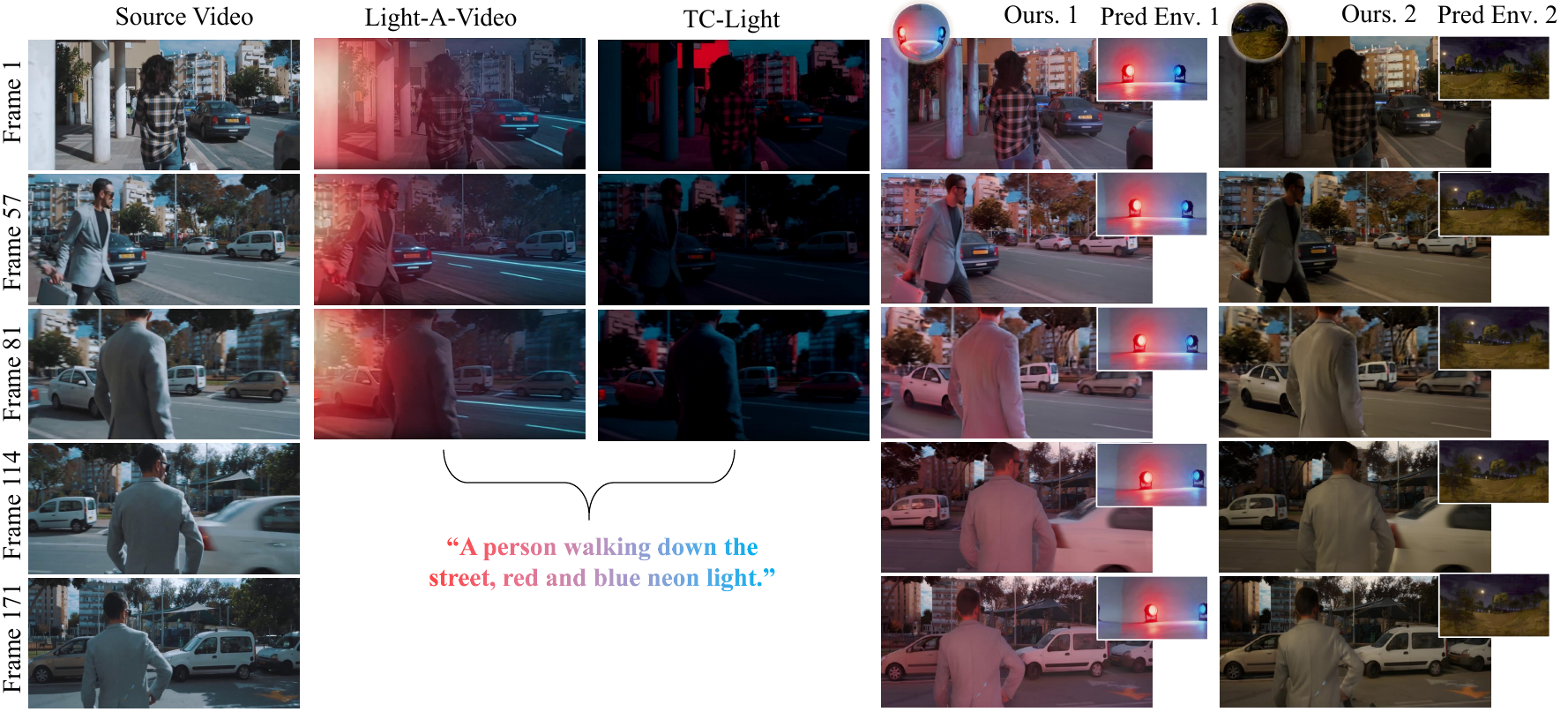}
  \caption{\textbf{Comparison of relighting results for long video sequences using different methods.} Light-A-Video and TC-Light process an entire 81-frame video in a single pass. Our approach divides a long video into multiple 57-frame segments, where the lighting conditions for each segment are derived from the lighting estimates of the preceding segment. In addition to the relighting results, our method also displays the corresponding predicted environment maps (converted from normalized log-intensity maps to LDR images for visualization).}
  \label{fig:eval_stream_relight}
\end{figure*}

In Figure~\ref{fig:eval_stream_relight}, we present additional results of our method for relighting long videos (along with comparisons to other methods).
Our method accurately perceives changes in the camera's viewpoint and correctly warp the environment map, thereby achieving temporally consistent lighting effects.

\subsection{Evaluation of environment video generation}

\begin{figure}[t]
  \centering
  \includegraphics[width=0.95\linewidth]{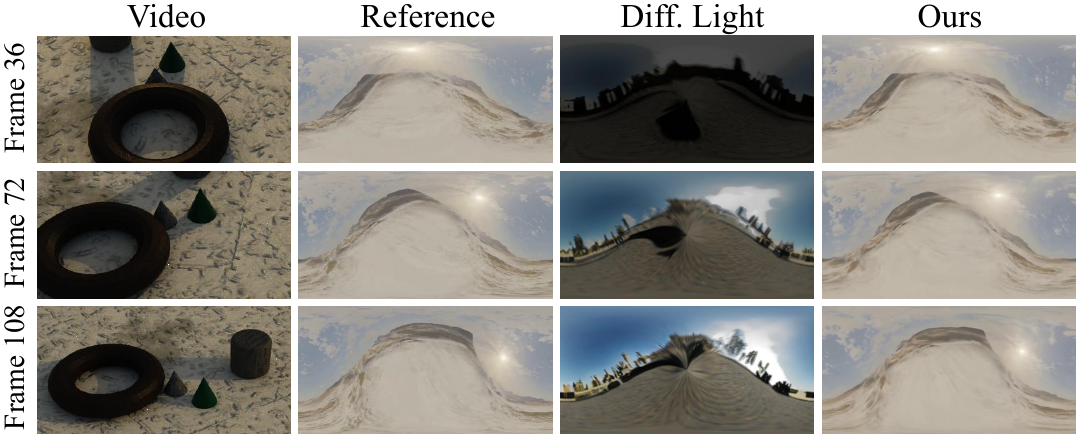}
  \caption{\textbf{Qualitative comparison of video lighting estimation on the synthetic dataset.} Given the environment map of the first frame, our method generates environment maps for every frame of the entire video. It produces smooth lighting deformations and accurately aligns with the camera viewpoint of each frame. We also provide the results of DiffusionLight~\cite{phongthawee2024diffusionlight}.}
  \label{fig:eval_light_est_3}
\end{figure}

As shown in Figure~\ref{fig:main}, our model can generate warped environment maps (i.e., environment video) , which can be viewed as a novel lighting estimation task that infers lighting for all frames based on the lighting of a single frame. In this section, we use common metrics to evaluate the accuracy of the generated warped environment maps. Accordingly, we provide results from several classic light estimation methods for reference, including StyleLight~\cite{wang2022stylelight} and DiffusionLight~\cite{phongthawee2024diffusionlight}. As shown in Figure~\ref{fig:eval_light_est_3}, the environment video produced by our method closely matches the reference. We also provide quantitative evaluation results in Table~\ref{tab:eval_light_est_2}, using metrics related to illumination direction, such as angular error, to evaluate our method's capability in detecting changes in camera pose. The results demonstrate the stability of our predictions over time, which is crucial for generating spatially accurate lighting in videos.

\begin{table}[]
\caption{\textbf{Directional angle error in video lighting estimation for sunlit scenes.} We use StyleLight~\cite{wang2022stylelight} and DiffusionLight~\cite{phongthawee2024diffusionlight} to estimate environment map frame-by-frame in videos, while our method generates the entire video's environment maps in a single pass. Note: The standard deviation (i.e., Std) here represents the average standard deviation of the directional errors across all videos.}
\label{tab:eval_light_est_2}
\resizebox{\linewidth}{!}{%
\begin{tabular}{l|ccc|ccc}
\toprule[1pt]
\multirow{3}{*}{Methods} & \multicolumn{3}{c|}{\multirow{2}{*}{\textbf{Angular Error top-5 ($\downarrow$)}}} & \multicolumn{3}{c}{\multirow{2}{*}{\textbf{Angular Error top-3 ($\downarrow$)}}} \\
                         & \multicolumn{3}{c|}{}                                                            & \multicolumn{3}{c}{}                                                       \\
                         & Mean                      & Median                   & Std                       & Mean                    & Median                 & Std                     \\ \midrule[0.5pt]
StyleLight               & 66.12                   & 62.62               & 24.11                  & 69.27                  & 63.81                & 26.23                        \\
DiffusionLight           & 54.88                   & 51.58               & 18.83                  & 55.18                  & 51.82                & 18.82                        \\
Ours                     & \textbf{20.35}                  & \textbf{7.96}                 & \textbf{14.14}                  & \textbf{20.69}                  & \textbf{7.76}                 & \textbf{14.31}                 \\ \bottomrule[1pt]
\end{tabular}
}
\end{table}

\subsection{Other applications}

\begin{figure}[t]
  \centering
  \includegraphics[width=0.45\textwidth]{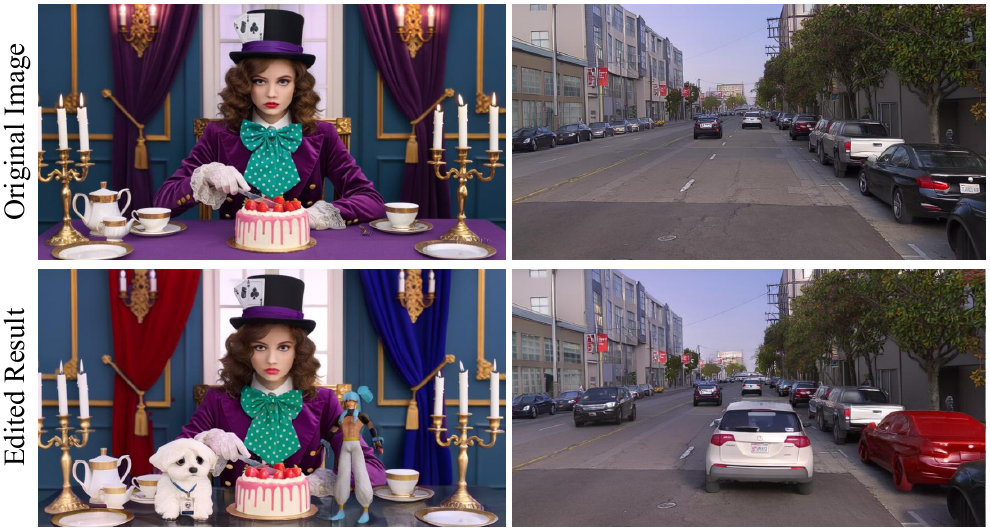}
  \caption{\textbf{Image editing application.} Left: Insert puppy and toy model into desktop scene; adjust base color of curtains and metallicness, roughness, and base color of tablecloth. Right: Insert vehicle into street scene; modify base color of entire vehicle on the right.}
   \label{fig:vis_app_1}
\end{figure}

\paragraph{Scene editing.}
Our method supports scene editing by utilizing scene rendering, as detailed in the supplemental material. In Figure~\ref{fig:vis_app_1}, we showcase object insertion and material editing, complete with realistic reflections and shadow effects.

\begin{figure}[t]
  \centering
  \includegraphics[width=\linewidth]{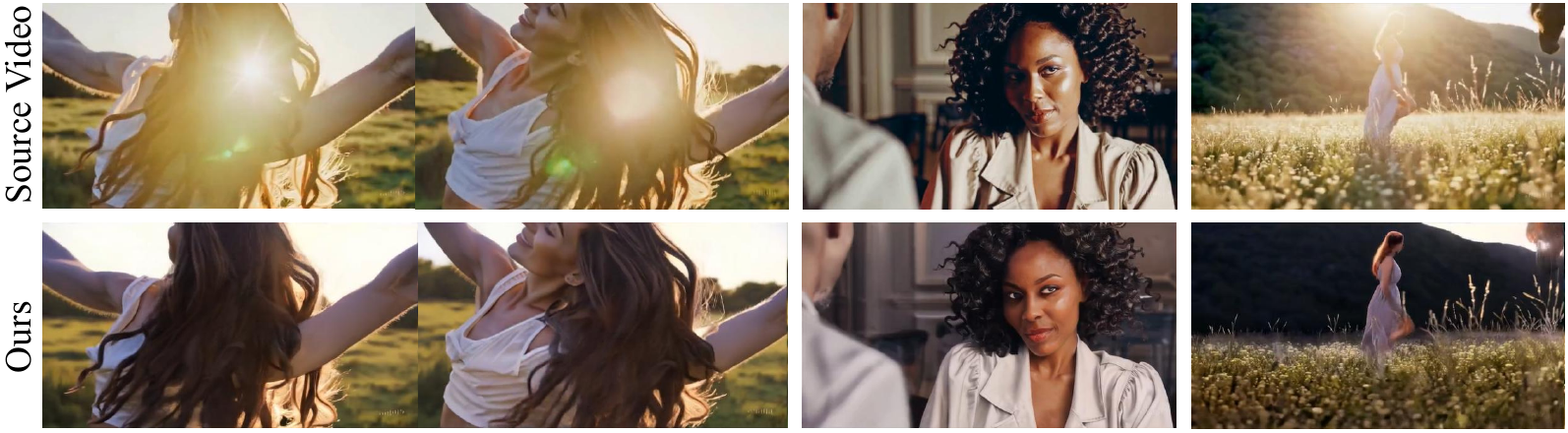}
  \caption{\textbf{Visualization results of video delighting.} Our method achieves realistic and natural lighting removal by resynthesizing video illumination through specific environmental map.}
  \label{fig:vis_app_2}
\end{figure}

\paragraph{Video delighting.}
 Our method also effectively removes specular highlights from the original video.
As shown in Figure~\ref{fig:vis_app_2}, our method accurately restores the original material properties in the delighted scene.
This is crucial for visual perception tasks~\cite{zheng2023steps} sensitive to specular artifacts, including 3D reconstruction and depth estimation.
Additional results are presented in supplemental material. 

\subsection{Ablation Study}
In this section, we conduct ablation studies on model architecture and training strategies to validate the effectiveness of the techniques proposed in this paper. See supplemental material for specific implementation details and further ablation studies. 

\begin{table}[]
\caption{\textbf{Ablation on different components.} Quantitative results of relighting on synthetic videos and MIT multi-illumination images. Both the raw image and the joint modeling of relighting with environmental video significantly enhance the model's relighting performance.}
\label{tab:eval_ablation_1}
\resizebox{\linewidth}{!}{%
\begin{tabular}{l|ccc|ccc}
\toprule[1pt]
\multirow{2}{*}{Methods} & \multicolumn{3}{c|}{Synthetic Video}                           & \multicolumn{3}{c}{MIT multi-illumination}                     \\
                         & PSNR ($\uparrow $) & SSIM ($\uparrow $) & LPIPS ($\downarrow$) & PSNR ($\uparrow $) & SSIM ($\uparrow $) & LPIPS ($\downarrow$) \\ \midrule[0.5pt]
w/o Env Video      & 17.45           & 0.703             & 0.284             & 21.21              & \textbf{0.851}              & \textbf{0.129}                \\
w/o Raw Image      &  20.84          &   0.715           &   0.291           & 18.73              & 0.743              & 0.225                \\

Full              &  \textbf{23.63}          &   \textbf{0.778}           &   \textbf{0.228}           & \textbf{21.49}              & 0.849              & 0.135                \\ \bottomrule[1pt]
\end{tabular}
}
\end{table}

\begin{figure}[t]
  \centering
  \includegraphics[width=0.48\textwidth]{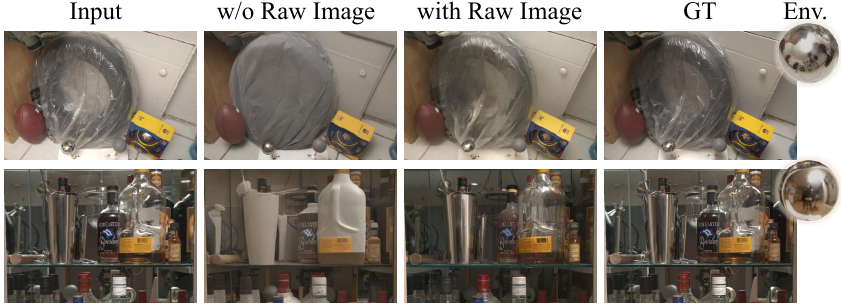}
  \caption{\textbf{Qualitative ablation of relighting.} The raw reference image significantly improves relighting quality on complex materials.}
   \label{fig:vis_ablation_3}
\end{figure}

\paragraph{Effectiveness of raw reference image.}
We present the quantitative ablation results of raw reference images in Table~\ref{tab:eval_ablation_1}.
The results demonstrate that introducing raw reference images significantly improves model performance.
Figure~\ref{fig:vis_ablation_3} presents a visual comparison.
As shown, the model without the raw reference image struggles to generate accurate physical transmission effects (notice the plastic bag and glass bottle in the scene).
This clearly demonstrates the effectiveness of the raw reference image: it corrects rendering errors caused by imperfect intrinsic decomposition and guides the model to learn realistic physical effects.

\paragraph{Effectiveness of joint generation.}
We also ablate the environment video generation branch in Table~\ref{tab:eval_ablation_1}.
The results demonstrate that, compared to the ablated model, our joint model achieves significant performance improvements on synthetic videos featuring scenes with substantial camera motion or dynamic lighting.
This fully demonstrates the benefits of environment video generation for camera-free video relighting.
By jointly generating relighting and environment video, the model effectively aligns the input environment map with each frame's camera viewpoint, thereby achieving spatially consistent video relighting.

\begin{figure}[t]
  \centering
  \includegraphics[width=0.8\linewidth]{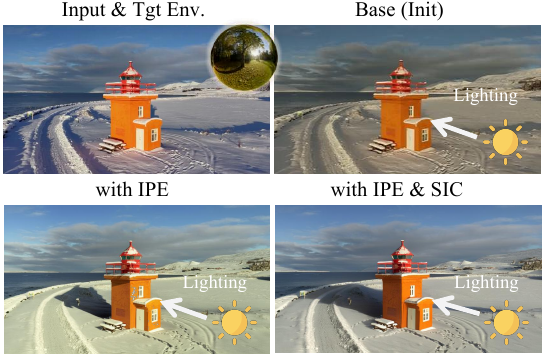}
  \caption{\textbf{Ablation on training strategies.} We mark the direction of peak illumination. IPE: Intrinsic Perception Enhancement. SIC: Self-supervised learning based on Illumination Consistency.}
  \label{fig:vis_training_stage}
\end{figure}

\paragraph{Training strategy validation.}
We conduct a comparative analysis of the design schemes for the three-stage training in Figure~\ref{fig:vis_training_stage}.
In fact, after the standard supervised training in the first stage, our initial model achieves a PSNR of 21 on the MIT multi-illumination benchmark, surpassing state-of-the-art models.
However, this model occasionally struggles with complex original lighting effects, as we lack training data on real-world scenes with multi-illumination conditions.
Since our ``Intrinsic Perception Enhancement'' strategy constructs a large number of pseudo-raw reference images with special lighting for real-world scenes, the model’s ability to decouple original lighting is significantly improved.
Furthermore, our self-supervised strategy enables closed-loop training under arbitrary lighting and scene, which further enhances the visual quality of relighting results.

\subsection{Limitations}
Although merging different intrinsic latents reduces computational overhead, the frame-dimensional concatenation-based control method still incurs substantial training costs.
Consequently, \ourmodel{} inevitably trades off resolution and frame rate, with a maximum of 57 frames achieved at $832\times 480$ resolution during training. Meanwhile, on the A800 GPU, generating a 57-frame video takes approximately 10 minutes.
\section{Conclusion}
\label{sec:conclution}

In this paper, we have presented \ourmodel{}, a novel video relighting framework that produces physically consistent, temporally stable results without requiring prior knowledge of camera pose. At the core of our framework is an RGB-intrinsic fusion renderer along with a joint generation formulation for the relit video and the environment video. This approach allows the model to incorporate real-world lighting effects while adhering to estimated physical constraints, resulting in realistic relighting outcomes. Additionally, the formulation eliminates the need for explicit pose estimation, enhancing practical flexibility. Furthermore, we design two complementary training strategies that effectively mitigate the scarcity of existing multi-light datasets and further improve the model's generalization in complex scenes. Extensive experiments confirm that \ourmodel{} outperforms state-of-the-art methods in generating physically consistent and temporally stable relighting results (e.g., shadows, reflections) with strong generalization. Additionally, its extensibility supports downstream tasks such as scene rendering and video illumination estimation, validating its potential as a universal video editing engine.


\bibliographystyle{ACM-Reference-Format}
\bibliography{bibliography}

@String{Computing = "Computing" }

@String{Computer = "{IEEE} Computer" }

@String{Springer = "Springer-Verlag" }

@inproceedings{liang2025diffusion,
  title={Diffusion Renderer: Neural Inverse and Forward Rendering with Video Diffusion Models},
  author={Liang, Ruofan and Gojcic, Zan and Ling, Huan and Munkberg, Jacob and Hasselgren, Jon and Lin, Chih-Hao and Gao, Jun and Keller, Alexander and Vijaykumar, Nandita and Fidler, Sanja and others},
  booktitle={Proceedings of the Computer Vision and Pattern Recognition Conference},
  pages={26069--26080},
  year={2025}
}

@article{fang2025relightvid,
  title={RelightVid: Temporal-consistent diffusion model for video relighting},
  author={Fang, Ye and Sun, Zeyi and Zhang, Shangzhan and Wu, Tong and Xu, Yinghao and Zhang, Pan and Wang, Jiaqi and Wetzstein, Gordon and Lin, Dahua},
  journal={arXiv preprint arXiv:2501.16330},
  year={2025}
}

@article{jin2024neural,
  title={Neural gaffer: Relighting any object via diffusion},
  author={Jin, Haian and Li, Yuan and Luan, Fujun and Xiangli, Yuanbo and Bi, Sai and Zhang, Kai and Xu, Zexiang and Sun, Jin and Snavely, Noah},
  journal={Advances in Neural Information Processing Systems},
  volume={37},
  pages={141129--141152},
  year={2024}
}

@article{fang2025v,
  title={V-RGBX: Video Editing with Accurate Controls over Intrinsic Properties},
  author={Fang, Ye and Wu, Tong and Deschaintre, Valentin and Ceylan, Duygu and Georgiev, Iliyan and Huang, Chun-Hao Paul and Hu, Yiwei and Chen, Xuelin and Wang, Tuanfeng Yang},
  journal={arXiv preprint arXiv:2512.11799},
  year={2025}
}

@article{beisswenger2025framediffuser,
  title={FrameDiffuser: G-Buffer-Conditioned Diffusion for Neural Forward Frame Rendering},
  author={Beisswenger, Ole and Dihlmann, Jan-Niklas and Lensch, Hendrik},
  journal={arXiv preprint arXiv:2512.16670},
  year={2025}
}

@article{ye2024stablenormal,
  title={Stablenormal: Reducing diffusion variance for stable and sharp normal},
  author={Ye, Chongjie and Qiu, Lingteng and Gu, Xiaodong and Zuo, Qi and Wu, Yushuang and Dong, Zilong and Bo, Liefeng and Xiu, Yuliang and Han, Xiaoguang},
  journal={ACM Transactions on Graphics (TOG)},
  volume={43},
  number={6},
  pages={1--18},
  year={2024},
  publisher={ACM New York, NY, USA}
}

@inproceedings{kocsisintrinsix,
  title={IntrinsiX: High-Quality PBR Generation using Image Priors},
  author={Kocsis, Peter and H{\"o}llein, Lukas and Nie{\ss}ner, Matthias},
  booktitle={The Thirty-ninth Annual Conference on Neural Information Processing Systems},
  year={2025}
}

@article{xi2025ctrlvdiff,
  title={CtrlVDiff: Controllable Video Generation via Unified Multimodal Video Diffusion},
  author={Xi, Dianbing and Wang, Jiepeng and Liang, Yuanzhi and Qiu, Xi and Liu, Jialun and Pan, Hao and Huo, Yuchi and Wang, Rui and Huang, Haibin and Zhang, Chi and others},
  journal={arXiv preprint arXiv:2511.21129},
  year={2025}
}

@inproceedings{liulight,
  title={Light-X: Generative 4D Video Rendering with Camera and Illumination Control},
  author={Liu, Tianqi and Chen, Zhaoxi and Huang, Zihao and Xu, Shaocong and Zhang, Saining and Ye, Chongjie and Li, Bohan and Cao, Zhiguo and Li, Wei and Zhao, Hao and others},
  booktitle={The Fourteenth International Conference on Learning Representations},
  year={2026}
}

@inproceedings{zhou2025light,
  title={Light-a-video: Training-free video relighting via progressive light fusion},
  author={Zhou, Yujie and Bu, Jiazi and Ling, Pengyang and Zhang, Pan and Wu, Tong and Huang, Qidong and Li, Jinsong and Dong, Xiaoyi and Zang, Yuhang and Cao, Yuhang and others},
  booktitle={Proceedings of the IEEE/CVF International Conference on Computer Vision},
  pages={13315--13325},
  year={2025}
}

@inproceedings{liuunilumos,
  title={UniLumos: Fast and Unified Image and Video Relighting with Physics-Plausible Feedback},
  author={Liu, Pengwei and Yuan, Hangjie and Dong, Bo and Xing, Jiazheng and Wang, Jinwang and Zhao, Rui and Chen, Weihua and Wang, Fan},
  booktitle={The Thirty-ninth Annual Conference on Neural Information Processing Systems},
  year={2025}
}

@article{bian2025relightmaster,
  title={Relightmaster: Precise video relighting with multi-plane light images},
  author={Bian, Weikang and Shi, Xiaoyu and Huang, Zhaoyang and Bai, Jianhong and Wang, Qinghe and Wang, Xintao and Wan, Pengfei and Gai, Kun and Li, Hongsheng},
  journal={arXiv preprint arXiv:2511.06271},
  year={2025}
}

@inproceedings{heunirelight,
  title={UniRelight: Learning Joint Decomposition and Synthesis for Video Relighting},
  author={He, Kai and Liang, Ruofan and Munkberg, Jacob and Hasselgren, Jon and Vijaykumar, Nandita and Keller, Alexander and Fidler, Sanja and Gilitschenski, Igor and Gojcic, Zan and Wang, Zian},
  booktitle={Advances in Neural Information Processing Systems},
  year={2025}
}

@inproceedings{chen2025invrgb+,
  title={InvRGB+ L: Inverse Rendering of Complex Scenes with Unified Color and LiDAR Reflectance Modeling},
  author={Chen, Xiaoxue and Chandaka, Bhargav and Lin, Chih-Hao and Zhang, Ya-Qin and Forsyth, David and Zhao, Hao and Wang, Shenlong},
  booktitle={Proceedings of the IEEE/CVF International Conference on Computer Vision},
  pages={27176--27186},
  year={2025}
}

@article{wang2025spatialvid,
  title={Spatialvid: A large-scale video dataset with spatial annotations},
  author={Wang, Jiahao and Yuan, Yufeng and Zheng, Rujie and Lin, Youtian and Gao, Jian and Chen, Lin-Zhuo and Bao, Yajie and Zhang, Yi and Zeng, Chang and Zhou, Yanxi and others},
  journal={arXiv preprint arXiv:2509.09676},
  year={2025}
}

@inproceedings{zeng2024rgb,
author = {Zeng, Zheng and Deschaintre, Valentin and Georgiev, Iliyan and Hold-Geoffroy, Yannick and Hu, Yiwei and Luan, Fujun and Yan, Ling-Qi and Ha\v{s}an, Milo\v{s}},
title = {RGB {$\leftrightarrow$} X: Image decomposition and synthesis using material- and lighting-aware diffusion models},
year = {2024},
isbn = {9798400705250},
publisher = {Association for Computing Machinery},
address = {New York, NY, USA},
url = {https://doi.org/10.1145/3641519.3657445},
doi = {10.1145/3641519.3657445},
booktitle = {ACM SIGGRAPH 2024 Conference Papers},
articleno = {75},
numpages = {11},
location = {Denver, CO, USA},
series = {SIGGRAPH '24}
}

@inproceedings{xu2025geometrycrafter,
  title={Geometrycrafter: Consistent geometry estimation for open-world videos with diffusion priors},
  author={Xu, Tian-Xing and Gao, Xiangjun and Hu, Wenbo and Li, Xiaoyu and Zhang, Song-Hai and Shan, Ying},
  booktitle={Proceedings of the IEEE/CVF International Conference on Computer Vision},
  pages={6632--6644},
  year={2025}
}

@article{careaga2023intrinsic,
  title={Intrinsic image decomposition via ordinal shading},
  author={Careaga, Chris and Aksoy, Ya{\u{g}}{\i}z},
  journal={ACM Transactions on Graphics},
  volume={43},
  number={1},
  pages={1--24},
  year={2023},
  publisher={ACM New York, NY, USA}
}

@article{ren2025mv,
  title={MV-CoLight: Efficient Object Compositing with Consistent Lighting and Shadow Generation},
  author={Ren, Kerui and Bai, Jiayang and Xu, Linning and Jiang, Lihan and Pang, Jiangmiao and Yu, Mulin and Dai, Bo},
  journal={arXiv preprint arXiv:2505.21483},
  year={2025}
}

@inproceedings{bin2025normalcrafter,
  title={Normalcrafter: Learning temporally consistent normals from video diffusion priors},
  author={Bin, Yanrui and Hu, Wenbo and Wang, Haoyuan and Chen, Xinya and Wang, Bing},
  booktitle={Proceedings of the IEEE/CVF International Conference on Computer Vision},
  pages={8330--8339},
  year={2025}
}

@inproceedings{liutc,
  title={TC-Light: Temporally Coherent Generative Rendering for Realistic World Transfer},
  author={Liu, Yang and Luo, Chuanchen and Tang, Zimo and Li, Yingyan and Ning, Yuanyong and Fan, Lue and Peng, Junran and Zhang, Zhaoxiang and others},
  booktitle={The Thirty-ninth Annual Conference on Neural Information Processing Systems},
  year={2025}
}

@inproceedings{phongthawee2024diffusionlight,
  title={Diffusionlight: Light probes for free by painting a chrome ball},
  author={Phongthawee, Pakkapon and Chinchuthakun, Worameth and Sinsunthithet, Nontaphat and Jampani, Varun and Raj, Amit and Khungurn, Pramook and Suwajanakorn, Supasorn},
  booktitle={Proceedings of the IEEE/CVF conference on computer vision and pattern recognition},
  pages={98--108},
  year={2024}
}

@article{rendering2015physically,
  title={Physically-based rendering},
  author={Rendering, Why Physically-Based},
  journal={Procedia IUTAM},
  volume={13},
  number={127-137},
  pages={3},
  year={2015},
  publisher={Elsevier}
}

@inproceedings{
    zhang2025scaling,
    title={Scaling In-the-Wild Training for Diffusion-based Illumination Harmonization and Editing by Imposing Consistent Light Transport},
    author={Lvmin Zhang and Anyi Rao and Maneesh Agrawala},
    booktitle={The Thirteenth International Conference on Learning Representations},
    year={2025},
    url={https://openreview.net/forum?id=u1cQYxRI1H}
}

@article{wan2025,
      title={Wan: Open and Advanced Large-Scale Video Generative Models}, 
      author={Team Wan and Ang Wang and Baole Ai and Bin Wen and Chaojie Mao and Chen-Wei Xie and Di Chen and Feiwu Yu and Haiming Zhao and Jianxiao Yang and Jianyuan Zeng and Jiayu Wang and Jingfeng Zhang and Jingren Zhou and Jinkai Wang and Jixuan Chen and Kai Zhu and Kang Zhao and Keyu Yan and Lianghua Huang and Mengyang Feng and Ningyi Zhang and Pandeng Li and Pingyu Wu and Ruihang Chu and Ruili Feng and Shiwei Zhang and Siyang Sun and Tao Fang and Tianxing Wang and Tianyi Gui and Tingyu Weng and Tong Shen and Wei Lin and Wei Wang and Wei Wang and Wenmeng Zhou and Wente Wang and Wenting Shen and Wenyuan Yu and Xianzhong Shi and Xiaoming Huang and Xin Xu and Yan Kou and Yangyu Lv and Yifei Li and Yijing Liu and Yiming Wang and Yingya Zhang and Yitong Huang and Yong Li and You Wu and Yu Liu and Yulin Pan and Yun Zheng and Yuntao Hong and Yupeng Shi and Yutong Feng and Zeyinzi Jiang and Zhen Han and Zhi-Fan Wu and Ziyu Liu},
      journal = {arXiv preprint arXiv:2503.20314},
      year={2025}
}

@inproceedings{yangcogvideox,
  title={CogVideoX: Text-to-Video Diffusion Models with An Expert Transformer},
  author={Yang, Zhuoyi and Teng, Jiayan and Zheng, Wendi and Ding, Ming and Huang, Shiyu and Xu, Jiazheng and Yang, Yuanming and Hong, Wenyi and Zhang, Xiaohan and Feng, Guanyu and others},
  booktitle={The Thirteenth International Conference on Learning Representations},
  year={2025}
}

@inproceedings{ling2024dl3dv,
  title={Dl3dv-10k: A large-scale scene dataset for deep learning-based 3d vision},
  author={Ling, Lu and Sheng, Yichen and Tu, Zhi and Zhao, Wentian and Xin, Cheng and Wan, Kun and Yu, Lantao and Guo, Qianyu and Yu, Zixun and Lu, Yawen and others},
  booktitle={Proceedings of the IEEE/CVF Conference on Computer Vision and Pattern Recognition},
  pages={22160--22169},
  year={2024}
}

@inproceedings{murmann2019multi,
  title={A multi-illumination dataset of indoor object appearance},
  author={Murmann, Lukas and Gharbi, Michael and Aittala, Miika and Durand, Fredo},
  booktitle={2019 IEEE international conference on computer vision (ICCV)},
  volume={2},
  number={8},
  year={2019}
}

@article{wei2025omnieraserremoveobjectseffects,
title={OmniEraser: Remove Objects and Their Effects in Images with Paired Video-Frame Data},
author={Runpu Wei and Zijin Yin and Shuo Zhang and Lanxiang Zhou and Xueyi Wang and Chao Ban and Tianwei Cao and Hao Sun and Zhongjiang He and Kongming Liang and Zhanyu Ma},
journal={arXiv preprint arXiv:2501.07397},
year={2025},
url={https://arxiv.org/abs/2501.07397},
}

@inproceedings{liu2024shadow,
  title={Shadow generation for composite image using diffusion model},
  author={Liu, Qingyang and You, Junqi and Wang, Jianting and Tao, Xinhao and Zhang, Bo and Niu, Li},
  booktitle={Proceedings of the IEEE/CVF Conference on Computer Vision and Pattern Recognition},
  pages={8121--8130},
  year={2024}
}

@article{qwen3,
    title={Qwen3 Technical Report}, 
    author={An Yang and Anfeng Li and Baosong Yang and Beichen Zhang and Binyuan Hui and Bo Zheng and Bowen Yu and Chang Gao and Chengen Huang and Chenxu Lv and Chujie Zheng and Dayiheng Liu and Fan Zhou and Fei Huang and Feng Hu and Hao Ge and Haoran Wei and Huan Lin and Jialong Tang and Jian Yang and Jianhong Tu and Jianwei Zhang and Jianxin Yang and Jiaxi Yang and Jing Zhou and Jingren Zhou and Junyang Lin and Kai Dang and Keqin Bao and Kexin Yang and Le Yu and Lianghao Deng and Mei Li and Mingfeng Xue and Mingze Li and Pei Zhang and Peng Wang and Qin Zhu and Rui Men and Ruize Gao and Shixuan Liu and Shuang Luo and Tianhao Li and Tianyi Tang and Wenbiao Yin and Xingzhang Ren and Xinyu Wang and Xinyu Zhang and Xuancheng Ren and Yang Fan and Yang Su and Yichang Zhang and Yinger Zhang and Yu Wan and Yuqiong Liu and Zekun Wang and Zeyu Cui and Zhenru Zhang and Zhipeng Zhou and Zihan Qiu},
    journal = {arXiv preprint arXiv:2505.09388},
    year={2025}
}

@inproceedings{wang2022stylelight,
  title={Stylelight: Hdr panorama generation for lighting estimation and editing},
  author={Wang, Guangcong and Yang, Yinuo and Loy, Chen Change and Liu, Ziwei},
  booktitle={European conference on computer vision},
  pages={477--492},
  year={2022},
  organization={Springer}
}

@inproceedings{vecchio2024matsynth,
  title={Matsynth: A modern pbr materials dataset},
  author={Vecchio, Giuseppe and Deschaintre, Valentin},
  booktitle={Proceedings of the IEEE/CVF Conference on Computer Vision and Pattern Recognition},
  pages={22109--22118},
  year={2024}
}

@article{li2025lightnormalsunifiedfeature,
      title={Light of Normals: Unified Feature Representation for Universal Photometric Stereo}, 
      author={Hong Li and Houyuan Chen and Chongjie Ye and Zhaoxi Chen and Bohan Li and Shaocong Xu and Xianda Guo and Xuhui Liu and Yikai Wang and Baochang Zhang and Satoshi Ikehata and Boxin Shi and Anyi Rao and Hao Zhao},
      journal={arXiv preprint arXiv:2506.18882},
      year={2025}
}

@inproceedings{chen2025uni,
  title={Uni-Renderer: Unifying Rendering and Inverse Rendering Via Dual Stream Diffusion},
  author={Chen, Zhifei and Xu, Tianshuo and Ge, Wenhang and Wu, Leyi and Yan, Dongyu and He, Jing and Wang, Luozhou and Zeng, Lu and Zhang, Shunsi and Chen, Ying-Cong},
  booktitle={Proceedings of the Computer Vision and Pattern Recognition Conference},
  pages={26504--26513},
  year={2025}
}

@inproceedings{helotus,
  title={Lotus: Diffusion-based Visual Foundation Model for High-quality Dense Prediction},
  author={He, Jing and Li, Haodong and Yin, Wei and Liang, Yixun and Li, Leheng and Zhou, Kaiqiang and Zhang, Hongbo and Liu, Bingbing and Chen, Ying-Cong},
  booktitle={The Thirteenth International Conference on Learning Representations},
  year={2025}
}

@inproceedings{bonneel2017intrinsic,
  title={Intrinsic decompositions for image editing},
  author={Bonneel, Nicolas and Kovacs, Balazs and Paris, Sylvain and Bala, Kavita},
  booktitle={Computer graphics forum},
  volume={36},
  number={2},
  pages={593--609},
  year={2017},
  organization={Wiley Online Library}
}

@inproceedings{shu2018deforming,
  title={Deforming autoencoders: Unsupervised disentangling of shape and appearance},
  author={Shu, Zhixin and Sahasrabudhe, Mihir and Guler, Riza Alp and Samaras, Dimitris and Paragios, Nikos and Kokkinos, Iasonas},
  booktitle={Proceedings of the European conference on computer vision (ECCV)},
  pages={650--665},
  year={2018}
}

@online{pexels2025,
  author = {Pexels},
  title = {Pexels Free Stock Media Platform},
  url = {https://www.pexels.com},
  year = {2025}
}

@misc{openai2024video,
  title={Video generation models as world simulators},
  author={OpenAI},
  year={2024}
}

@inproceedings{xiao2021pandaset,
  title={Pandaset: Advanced sensor suite dataset for autonomous driving},
  author={Xiao, Pengchuan and Shao, Zhenlei and Hao, Steven and Zhang, Zishuo and Chai, Xiaolin and Jiao, Judy and Li, Zesong and Wu, Jian and Sun, Kai and Jiang, Kun and others},
  booktitle={2021 IEEE international intelligent transportation systems conference (ITSC)},
  pages={3095--3101},
  year={2021},
  organization={IEEE}
}

@inproceedings{walke2023bridgedata,
  title={Bridgedata v2: A dataset for robot learning at scale},
  author={Walke, Homer Rich and Black, Kevin and Zhao, Tony Z and Vuong, Quan and Zheng, Chongyi and Hansen-Estruch, Philippe and He, Andre Wang and Myers, Vivek and Kim, Moo Jin and Du, Max and others},
  booktitle={Conference on Robot Learning},
  pages={1723--1736},
  year={2023},
  organization={PMLR}
}

@article{ren2024grounded,
  title={Grounded sam: Assembling open-world models for diverse visual tasks},
  author={Ren, Tianhe and Liu, Shilong and Zeng, Ailing and Lin, Jing and Li, Kunchang and Cao, He and Chen, Jiayu and Huang, Xinyu and Chen, Yukang and Yan, Feng and others},
  journal={arXiv preprint arXiv:2401.14159},
  year={2024}
}

@inproceedings{magar2025lightlab,
  title={Lightlab: Controlling light sources in images with diffusion models},
  author={Magar, Nadav and Hertz, Amir and Tabellion, Eric and Pritch, Yael and Rav-Acha, Alex and Shamir, Ariel and Hoshen, Yedid},
  booktitle={Proceedings of the Special Interest Group on Computer Graphics and Interactive Techniques Conference Conference Papers},
  pages={1--11},
  year={2025}
}

@inproceedings{careaga2025physically,
  title={Physically controllable relighting of photographs},
  author={Careaga, Chris and Aksoy, Ya{\u{g}}{\i}z},
  booktitle={Proceedings of the Special Interest Group on Computer Graphics and Interactive Techniques Conference Conference Papers},
  pages={1--10},
  year={2025}
}

@article{zheng2023steps,
  title={Steps: Joint self-supervised nighttime image enhancement and depth estimation},
  author={Zheng, Yupeng and Zhong, Chengliang and Li, Pengfei and Gao, Huan-ang and Zheng, Yuhang and Jin, Bu and Wang, Ling and Zhao, Hao and Zhou, Guyue and Zhang, Qichao and others},
  journal={arXiv preprint arXiv:2302.01334},
  year={2023}
}

@inproceedings{chen2025physgen3d,
  title={Physgen3d: Crafting a miniature interactive world from a single image},
  author={Chen, Boyuan and Jiang, Hanxiao and Liu, Shaowei and Gupta, Saurabh and Li, Yunzhu and Zhao, Hao and Wang, Shenlong},
  booktitle={Proceedings of the IEEE/CVF Conference on Computer Vision and Pattern Recognition},
  pages={6178--6189},
  year={2025}
}

\clearpage
\setcounter{page}{1}

\appendix   
\setcounter{table}{0}   
\setcounter{section}{0}
\setcounter{equation}{0}

\renewcommand{\thetable}{A\arabic{table}}
\renewcommand{\thefigure}{A\arabic{figure}}
\renewcommand{\theequation}{A\arabic{equation}}


\section{Appendix}
\label{appendix}

\subsection{Data generation strategy}
\label{sec:data_curation}

Our training data comprises a substantial synthetic dataset alongside auto-labeled real-world datasets, specifically as follows:

\parahead{Synthetic dataset.}
We produce a large number of rendered videos through a data synthesis workflow, complete with their base colors, roughness, metallicness, normal maps, depth maps, environment maps, and camera trajectories. Specifically, we first collect 5700 high-quality PBR material maps and 2241 HDR environment maps from public resources~\cite{vecchio2024matsynth,li2025lightnormalsunifiedfeature}. For each scene, we initially place a plane as well as up to 12 primitives (such as cubes, cones, and cylinders), with collision detection applied to avoid object intersections. Subsequently, we select PBR material maps randomly to texture both the plane and the primitives. Finally, we generate three types of videos with random motion patterns, namely: 1) Camera rotation with fixed lighting; 2) Lighting rotation with fixed camera; 3) Simultaneous rotation of both the camera and lighting. Each scene is rendered under at least two random lighting conditions for the same motion pattern. In total, we generate 8,000 videos, each consisting of 120 frames at a resolution of $512\times 512$. 

\parahead{Real-world dataset.}
We collected a large number of video clips and images from real-world datasets, including DL3DV~\cite{ling2024dl3dv}, SpatialVID-HQ~\cite{wang2025spatialvid}, MIT multi-illumination~\cite{murmann2019multi}, RemovalBench~\cite{wei2025omnieraserremoveobjectseffects}, and SOBAv2~\cite{liu2024shadow}. Specifically, we used a Vision Language Model (VLM)~\cite{qwen3} to filter these datasets (excluding MIT multi-illumination), removing data with blurred frames or significant shadows from objects outside the frame. 
Through this filtering process, we selected 13,809 video clips (57 frames) and 792 images.
We then generated pseudo ground-truth G-buffers for the above data using Diffusion Renderer's inverse renderer. 
For MIT multi-illumination, we exported environment maps based on the dataset's included reflective chrome sphere screenshots. 
For other datasets, we employed the VLM to determine the camera perspective of all data samples, and applied DiffusionLight~\cite{phongthawee2024diffusionlight} exclusively to annotate environment maps for images with horizontal perspectives, while manually filtering out results with obvious errors. Because we find that the DiffusionLight tends to yield significant estimation errors for images captured from non-horizontal perspectives. 
Ultimately, we annotated environmental maps for 8,278 videos and 206 images, performing frame-by-frame alignment based on camera trajectories.
These datasets with varying levels of completeness, combined with image data from MIT multi-illumination, collectively form a real-world dataset that significantly enriches our training samples for realistic scenarios.

\subsection{Initial training}
\label{sec:std_training}
Considering the significant differences in illumination distribution between synthetic and real-world datasets, we dynamically adjust our training mode across different periods. Specifically, we initially train the model exclusively on synthetic data to learn foundational rendering. Subsequently, we freeze the cross-attention module and train on the full dataset to enhance generalization while preserving adaptability to varying lighting conditions. During training, we set the latent $\mathbf{z}^\mathbf{I}$ to zero with a probability of 0.3 to simulate a pure rendering task. Following UniRelight~\cite{heunirelight}, for the sets of real-world datasets with and without environment maps, the denoising targets become $\hat{\mathbf{z}}^\mathbf{s}(\theta ),\hat{\mathbf{z}}^\mathbf{E_{log}}(\theta ) = \mathbf{f}_\theta  ([\mathbf{z}^\mathbf{I},\mathbf{z}_{\tau }^\mathbf{s},\mathbf{z}_{\tau }^\mathbf{E_{log}},\mathbf{z} ^{\left \{\mathbf{a},\mathbf{d},\mathbf{m}\right \}},\mathbf{z} ^{\left \{\mathbf{n},\mathbf{r}\right \}}+\mathbf{c_E}];\mathbf{c}_\mathbf{E}^{\mathrm{cross}},\tau )$ and $\hat{\mathbf{z}}^\mathbf{t}(\theta ) = \mathbf{f}_\theta  ([\mathbf{z}^\mathbf{I},\mathbf{z}_{\tau }^\mathbf{t},\mathbf{z}_{\tau }^\mathbf{0},\mathbf{z} ^{\left \{\mathbf{a},\mathbf{d},\mathbf{m}\right \}},\mathbf{z} ^{\left \{\mathbf{n},\mathbf{r}\right \}}+\mathbf{0}];\mathbf{0},\tau )$ respectively, enabling training on real-world scenes with single lighting conditions.

\begin{figure*}[t]
  \centering
  \includegraphics[width=1\textwidth]{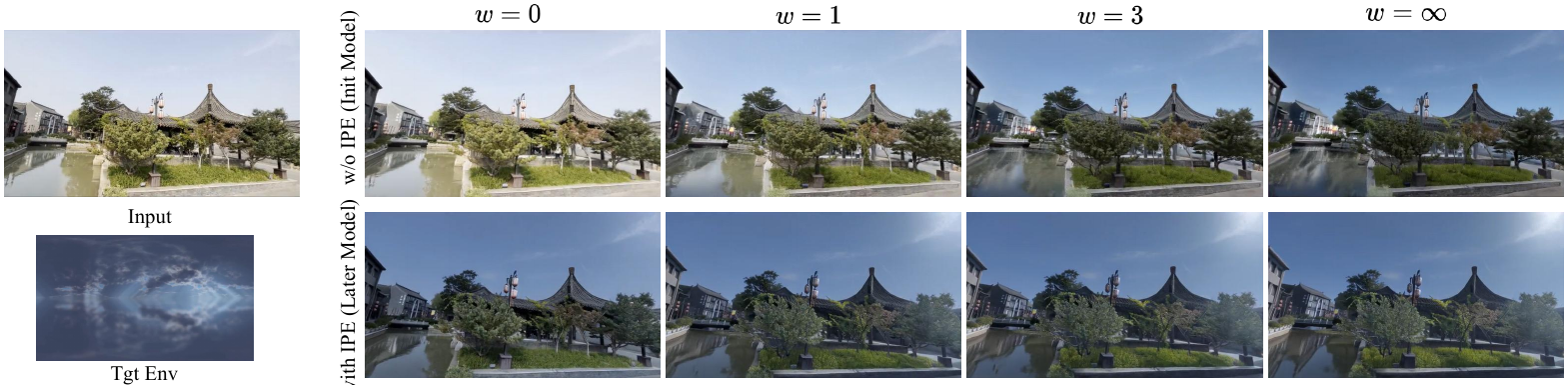}
  \caption{\textbf{Example of latent space interpolation between relighting and rendering results.} We demonstrate model interpolation outcomes before and after applying the Intrinsic Perception Enhancement strategy. Note: When $w=0$, the interpolated result is fully equivalent to the relighting result; when $w=\infty $, it is fully equivalent to the rendering result. This visualization not only illustrates the interpolation process but also validates the effectiveness of the Intrinsic Perception Enhancement strategy.}
   \label{fig:lantent_light}
\end{figure*}

\subsection{Additional Details on Intrinsic Perception Enhancement}
\label{sec:detail_IPE}
In this section, we focus on the details of multi-illumination data generation in Intrinsic Perception Enhancement.
In fact, this approach was inspired by our observations of the initial model's generated results.
As introduced in our discussion of the \ourmodel{} architecture's flexibility, our model supports both relighting (w/ $\mathbf{z}^\mathbf{I}$) and rendering (w/o $\mathbf{z}^\mathbf{I}$) tasks.
The distinction in their specific inference processes lies solely in whether the raw reference image is input.
Ideally, these two inference modes should produce identical results for the same scene.
But that's not the case.
We found that for relighting mode, the initial model has a certain probability of extracting the original lighting from the raw reference image.
In other words, while the information in the raw reference image significantly shapes the high-quality details of the relit video, it may also cause the result to deviate from the target lighting.
We illustrate an example in the first row and first column of Figure~\ref{fig:lantent_light}.
Overall, relighting results exhibit high quality with occasional lighting anomalies, while rendering results feature reasonable lighting but lack visual realism.
Therefore, we attempt to fuse both outputs to achieve stable, high-quality data augmentation with consistent lighting. During the implementation of Intrinsic Perception Enhancement, we generated multi-illumination data multiple times using the optimal model at each corresponding time point to continuously improve the quality of generated data.
For each multi-illumination data generation, we adjusted the appropriate parameter $w$ for the corresponding model.
Figure~\ref{fig:lantent_light} simultaneously displays the multi-illumination data generated by the initial model and the later model.
It is evident that the quality of our generated data has achieved substantial improvement.
This high-quality data enhances our model's ability to decouple original illumination, thereby improving relighting performance.

\subsubsection*{Potential Discussion: Why not use existing relighting models to generate multi-illumination data?}
According to the IPE strategy, the generated data serves as (pseudo) original reference images. These should exhibit accurate and physically realistic lighting effects, which existing open-source relighting models cannot achieve. As demonstrated in Section~\ref{sec:video_relight}, even advanced open-source methods still fail to produce realistic relighting results that adhere to material properties.

\subsection{Experimental details}
\label{sec:experiments_detail}

\parahead{Training Details.}
We adopt Wan2.1-T2V-1.3B~\cite{wan2025} as the base model and achieve our \ourmodel{} by fine-tuning its components. As mentioned in Section~\ref{sec:traing_strategy}, our training is divided into three stages. In the first stage, we first train the model for 10,000 iterations using synthetic data only, then train it for 20,000 iterations on the full dataset to obtain the initial model. In the second stage, we use the initial model to generate 8 pseudo-realistic images with different illuminations for each scenario in the real-world dataset, training the model for 5,000 iterations. In the third stage, we implement the SIC strategy with a probability of 0.1, training the model for a final 5,000 iterations. All training is conducted on 8 A800 GPUs with a batch size of 16, resolution of $832\times 480$, and AdamW optimizer with a learning rate of 1e-5. Total training takes about 7 days. We only fix the training video length to 17 during the synthetic data training in the first stage. In subsequent processes, the training video length is incrementally increased cyclically (from 1 to 57, following the 8n+1 pattern) to ensure the model's generalization capability across different frame lengths.

\parahead{Baselines.}
For video relighting, we compare \ourmodel{} against multiple advanced video relighting methods, including UniRelight~\cite{heunirelight}(as of now, it has not been open-sourced, so we directly replicate the results from the paper report), Cosmos-Diffusion Renderer~\cite{liang2025diffusion}, Light-A-Video~\cite{zhou2025light}, TC-Light~\cite{liutc}, and advanced image relighting method NeuralGaffer~\cite{jin2024neural}. For scene rendering, we compare our approach with two representative neural rendering methods RGBX~\cite{zeng2024rgb} and Cosmos-Diffusion Renderer. For environment light estimation, we compare with the image lighting estimation methods DiffusionLight~\cite{phongthawee2024diffusionlight} and StyleLight~\cite{wang2022stylelight}.

\parahead{Dataset.}
We curate test datasets through multiple channels for evaluating various tasks. First, we have created a high-quality synthetic test set, comprising 1,000 high-motion videos, each consisting of 120 frames (covering the three ``camera-light'' motion patterns mentioned in Section~\ref{sec:data_curation}). Meanwhile, we employ the MIT multi-illumination test set~\cite{murmann2019multi}, comprising 30 high-quality scenes across 25 lighting configurations. To ensure fair comparison, the evaluation methodology of relighting on this dataset is identical to that used in UniRelight: images under the i-th lighting condition are paired with those under the (i+12)-th lighting condition to form test pairs. Additionally, we have collected a series of videos from diverse domains, encompassing scenarios such as portraits, nature, roadways, and robotics, to evaluate the method's generalization in the real world.
Specifically, we collected 277 high-quality videos from Pexels~\cite{pexels2025} and Sora~\cite{openai2024video}, covering subjects such as humans, animals, and objects, and including various camera movements and object motions.
Beyond that, we also collected 100 representative videos each from PandaSet~\cite{xiao2021pandaset} and Bridgev2~\cite{walke2023bridgedata} for evaluation in embodied and autonomous driving domains.
These videos are completely unrelated to our training data. 

\parahead{Evaluation metrics.}
  Due to differences in dataset composition, we conduct distinct evaluations on different test sets.
  1) We evaluate the performance of relighting, rendering, and lighting estimation simultaneously on the synthetic test set and the MIT multi-illumination test set. (i) For relighting and rendering, we employ PSNR, SSIM, and LPIPS as evaluation metrics to frame-by-frame assess the visual fidelity between generated results and ground truth. (ii) For illumination estimation, we report angular error in degrees in scenes with concentrated sunlight.

2) For real-world videos across various domains under other single-lighting conditions, we evaluate the motion preservation and material consistency of relit results using existing pre-trained models, and assess the physical consistency of relit effects through user studies. (i) Specifically, we employ RAFT to estimate optical flow for both the source video and the relit video. The motion preservation scores for each method are evaluated by calculating the optical flow differences. (ii) Then, we evaluate material consistency between the source video and the generated video by calculating the average CLIP score (CLIP-MC) and average DINOv3 score (DINO-MC) for corresponding frames according to~\ref{eq:mc}, where higher scores indicate better consistency. It is worth noting that our CLIP scores are computed between the source video and the generated video, rather than between consecutive frames as in previous methods~\cite{zhou2025light,liulight} (which are used to measure video smoothness). To validate these metrics, we produce paired datasets with identical layouts but varying lighting/materials, and compute DINOv3 and CLIP similarity. Both metrics exhibit invariance (\(\ge0.94\)) to lighting variations and sensitivity (\(\le0.89\)) to object material differences. (iii) Our user study focuses on whether generated results are physically plausible, which is crucial for fields like simulation that demand high realism. Specifically, the evaluation involves visual realism (VR), physical consistency (PC) and lighting alignment(LA).

\begin{equation}
\text{MC} = 1 - \frac{1}{2N}\sum_{i=1}^{N} (1 - \cos(\mathbf{f}_i^{\text{src}}, \mathbf{f}_i^{\text{gen}})) \in [0, 1]
\label{eq:mc}
\end{equation}

\subsection{User Study}
\label{sec:user_study}
\begin{figure}[t]
  \centering
  \includegraphics[width=0.48\textwidth]{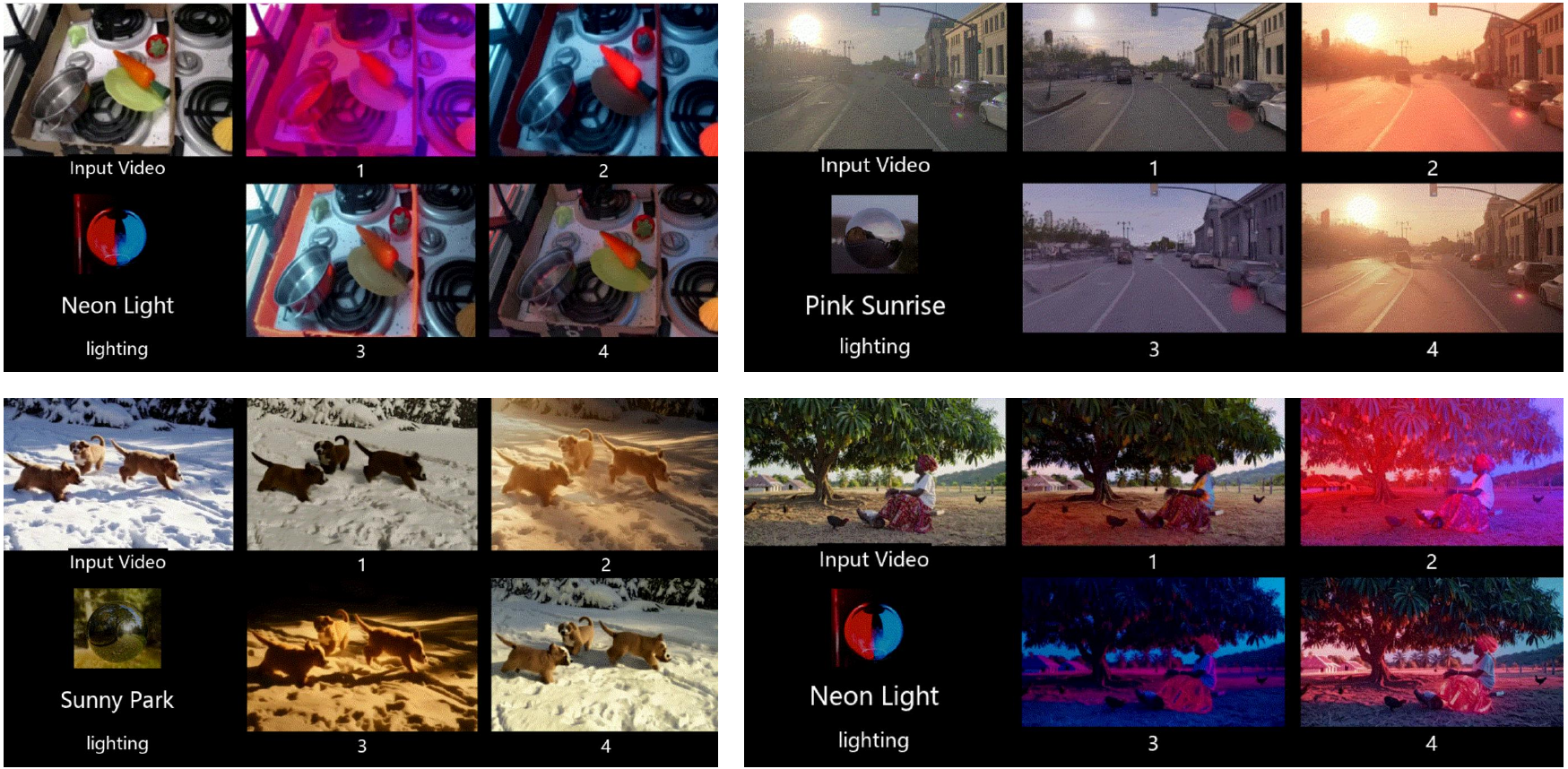}
  \caption{\textbf{The visual interface for our user research.} Participants simultaneously observed the input video, target lighting (text or environment map) and the results of four methods (randomly shuffled) displayed side-by-side. They evaluated each set of results based on three criteria by selecting methods that clearly failed.}
   \label{fig:user_study}
\end{figure}

Figure~\ref{fig:user_study} illustrates the interface used for our user study. Participants evaluated the results based on three questions:
(i) Which result is inconsistent with the input video's realism (e.g., anomalous glowing or artifacts)?
(ii) Which result fails to modify original shadows or metallic highlights?
(iii) Which result exhibits lighting inconsistent with the target condition shown in the bottom-left (defined by text description or environment map)?
For each question, participants could select between 0 and 2 options. In total, we collected
responses from 37 participants. Each participant was required to complete 10 sets of comparisons. We recorded instances where our method outperformed the baselines and vice versa. It is worth noting that in cases of a tie, we administered the survey repeatedly until a clear judgment was reached. Finally, we calculated the ratio of our method outperforming each baseline as the final metric.

\subsection{Evaluation of forward rendering}

Besides video relighting, \ourmodel{} also supports forward rendering. To validate our model, we compare the neural rendering performance of different methods in Table~\ref{tab:eval_render_all}. 
Note that our synthetic video dataset includes dynamic cameras, dynamic lighting, and combinations of both. These three motion patterns present fundamentally different challenges for our method, as its lighting conditions originate from the initial viewpoint. Both dynamic cameras and dynamic lighting introduce additional complexity for our approach. In contrast, this poses no significant theoretical difference for Diffusion Renderer, which defines environment maps frame-by-frame. Nevertheless, our approach achieves optimal performance, showcasing its robustness across diverse dynamic scenes.

\begin{figure}[t]
  \centering
  \includegraphics[width=0.5\textwidth]{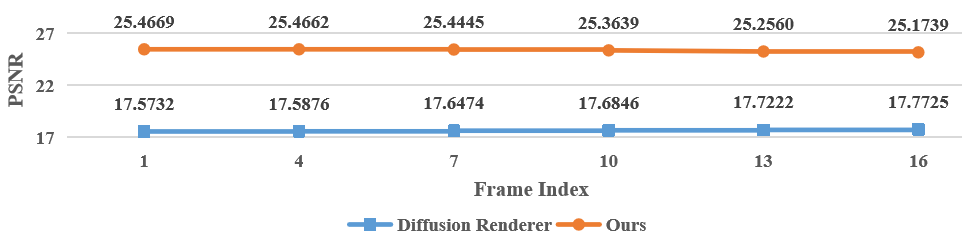}
  \caption{\textbf{Error distribution across different frames.} We compute the average PSNR for each frame across 200 synthetic videos. Our results exhibit high stability over time, comparable to methods that condition on per-frame environment maps.}
   \label{fig:eval_render_1}
\end{figure}

We visualize the error distribution of each method across different frames for video rendering task, as shown in Figure~\ref{fig:eval_render_1}.
Our method and Diffusion Renderer both demonstrate relatively stable rendering performance across different frames.
Despite only being fed the environment map from the initial viewpoint, our model robustly perceives camera viewpoint changes, accurately propagating the lighting from the initial viewpoint to all viewpoints.

\begin{table*}[]
\caption{\textbf{Quantitative comparison of neural rendering on the synthetic dataset.} The D. denotes dynamic, while the S. denotes static.}
\label{tab:eval_render_all}
\resizebox{\linewidth}{!}{%
\begin{tabular}{l|ccc|ccccccccc}
\toprule[1pt]
\multirow{3}{*}{Methods} & \multicolumn{3}{c|}{\multirow{2}{*}{\textit{Synthetic Image}}} & \multicolumn{9}{c}{\textit{Synthetic Video}}                                                                                                                                                                                 \\ \cmidrule{5-13} 
                         & \multicolumn{3}{c|}{}                                                       & \multicolumn{3}{c|}{S.Lighting - D.Camera}                               & \multicolumn{3}{c|}{D.Lighting - S.Camera}                               & \multicolumn{3}{c}{D.Lighting - D.Camera}          \\ \cmidrule{2-13} 
                         & PSNR ($\uparrow $)      & SSIM ($\uparrow $)     & LPIPS ($\downarrow$)     & PSNR ($\uparrow $) & SSIM ($\uparrow $) & \multicolumn{1}{c|}{LPIPS ($\downarrow$)} & PSNR ($\uparrow $) & SSIM ($\uparrow $) & \multicolumn{1}{c|}{LPIPS ($\downarrow$)} & PSNR ($\uparrow $) & SSIM ($\uparrow $) & LPIPS ($\downarrow$) \\ \midrule[0.5pt]
RGBX~\cite{zeng2024rgb}                     & 11.97                 & 0.503              & 0.429            & -                & -                & \multicolumn{1}{c|}{-}                    & -                  & -                  & \multicolumn{1}{c|}{-}                    & -                  & -                       & -                         \\
Diffusion Renderer~\cite{liang2025diffusion}        & 18.71                 & 0.692              & 0.256            & 18.09            & 0.676            & \multicolumn{1}{c|}{0.265}                & 17.46              & 0.662              & \multicolumn{1}{c|}{0.256}                & 17.72              & 0.666                   & 0.274                     \\
Ours                     & \textbf{24.31}        & \textbf{0.760}     & \textbf{0.197}   & \textbf{25.09}   & \textbf{0.792}   & \multicolumn{1}{c|}{\textbf{0.216}}       & \textbf{25.13}     & \textbf{0.810}     & \multicolumn{1}{c|}{\textbf{0.201}}       & \textbf{25.59}     & \textbf{0.802}          & \textbf{0.215}            \\ \bottomrule[1pt]
\end{tabular}
}
\end{table*}

\subsection{Supplement to ablation studies}
\label{sec:additional_ablation}

\subsubsection*{Experimental settings.}
For the ablation studies on model architecture, we first train the model for 10,000 iterations using only synthetic data, followed by training it for 20,000 iterations on the full dataset. Throughout training, we employ standard supervised learning without incorporating the two training strategies proposed in this paper. For the ablation studies on training strategies, we additionally train the models from the architecture ablation, applying each strategy sequentially for 5,000 iterations. All training runs on a single A800 GPU with a batch size of 4, consistently fixing the training video length to 17 frames to reduce computational overhead.
Note that to ensure fairness and rigor, we train each ablation model starting from the base model Wan2.1-T2V-1.3B, using the same number of training steps. These models are unrelated to our final model.

\subsubsection*{Why use G-buffer latent group-wise addition?}
We use group-wise addition to preserve semantic separability within each group. This strategy is also used in UniRelight. In this section, we perform the ablation of the operation and present the results in Table~\ref{tab:eval_latent_addition}.
Experiments demonstrate that our group-wise addition performs comparably to frame-concat while consuming ~25\% fewer resources.

\begin{table*}[]
\caption{\textbf{Ablation of G-buffer latent group-wise addition.} We compare the performance of the Frame-concat method and our method on two datasets: Synthetic Video and MIT multi-illumination. Performance metrics include PSNR, SSIM, LPIPS, and GPU memory usage for both training and inference.}
\label{tab:eval_latent_addition}
\resizebox{\linewidth}{!}{%
\begin{tabular}{l|ccc|ccc|c|c}
\toprule[1pt]
\multirow{2}{*}{\textbf{Method}} & \multicolumn{3}{c|}{\textbf{Synthetic Video}} & \multicolumn{3}{c|}{\textbf{MIT multi-illumination}} & \textbf{Training\(_{57frames}\)} & \textbf{Inference\(_{57frames}\)} \\ \cmidrule[0.5pt](lr){2-7}
                                & PSNR\(\uparrow\) & SSIM\(\uparrow\) & LPIPS\(\downarrow\) & PSNR\(\uparrow\) & SSIM\(\uparrow\) & LPIPS\(\downarrow\) & GPU\_memory\(_{train}\downarrow\)(MiB) & GPU\_memory\(_{inference}\downarrow\)(MiB) \\ \midrule[0.5pt]
Frame-concat                    & 23.55            & \textbf{0.789}    & \textbf{0.214}       & 21.21            & \textbf{0.851}    & 0.137                & 69678                        & 33154                        \\
Ours                             & \textbf{23.63}   & 0.778             & 0.228                & \textbf{21.49}   & 0.849             & \textbf{0.135}       & \textbf{51990}               & \textbf{26840}               \\ \bottomrule[1pt]
\end{tabular}
}
\end{table*}

\subsubsection*{Why choose dual-input light conditions?}
\label{ablation:light}
\begin{figure}[t]
  \centering
  \includegraphics[width=0.45\textwidth]{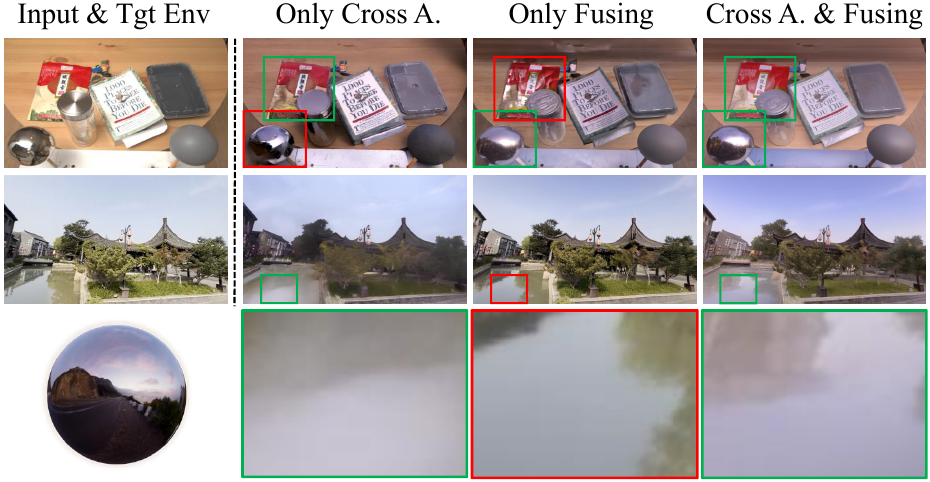}
  \caption{\textbf{Qualitative comparison under different light conditions.} Only Cross A.: Input light conditions via cross-attention. Only Fusing: By fusing with scene property latents to input light conditions. The dual-path lighting control method achieves the most balanced performance in terms of reflection quality and color temperature.}
   \label{fig:vis_ablation_2}
\end{figure}

We compare the results of relighting under different lighting control implementations, as shown in Figure~\ref{fig:vis_ablation_2}.
We find that compared to models that only inject lighting conditions via cross-attention, models that solely fuse environment light features into scene-intrinsic latent achieve better visual quality in overall detail across natural scenes.
However, this does not imply an architectural advantage.
As shown in the red box, in natural scenes, this model tends to directly replicate content from the raw reference image, often leading to degradation in relighting tasks.
In contrast, methods that inject lighting conditions solely through cross-attention generate color temperatures closer to the target lighting, but lack reflective details.
We speculate this is because our base model is a T2V model, which transmits text prompt information to the video generation model via cross-attention. However, textual information inherently possesses a natural sparsity compared to image data, making it challenging to describe fine-grained spatial structural details.

In other words, relying solely on cross-attention to input lighting conditions fails to accurately convey texture details from environment maps, resulting in degraded reflection effects.
In contrast, the duplicate light control approach achieves the best overall performance and is therefore adopted in our final architecture.

\subsection{Scene editing workflow}
\label{sec:editing_pipeline}
\begin{figure*}[t]
  \centering
  \includegraphics[width=\textwidth]{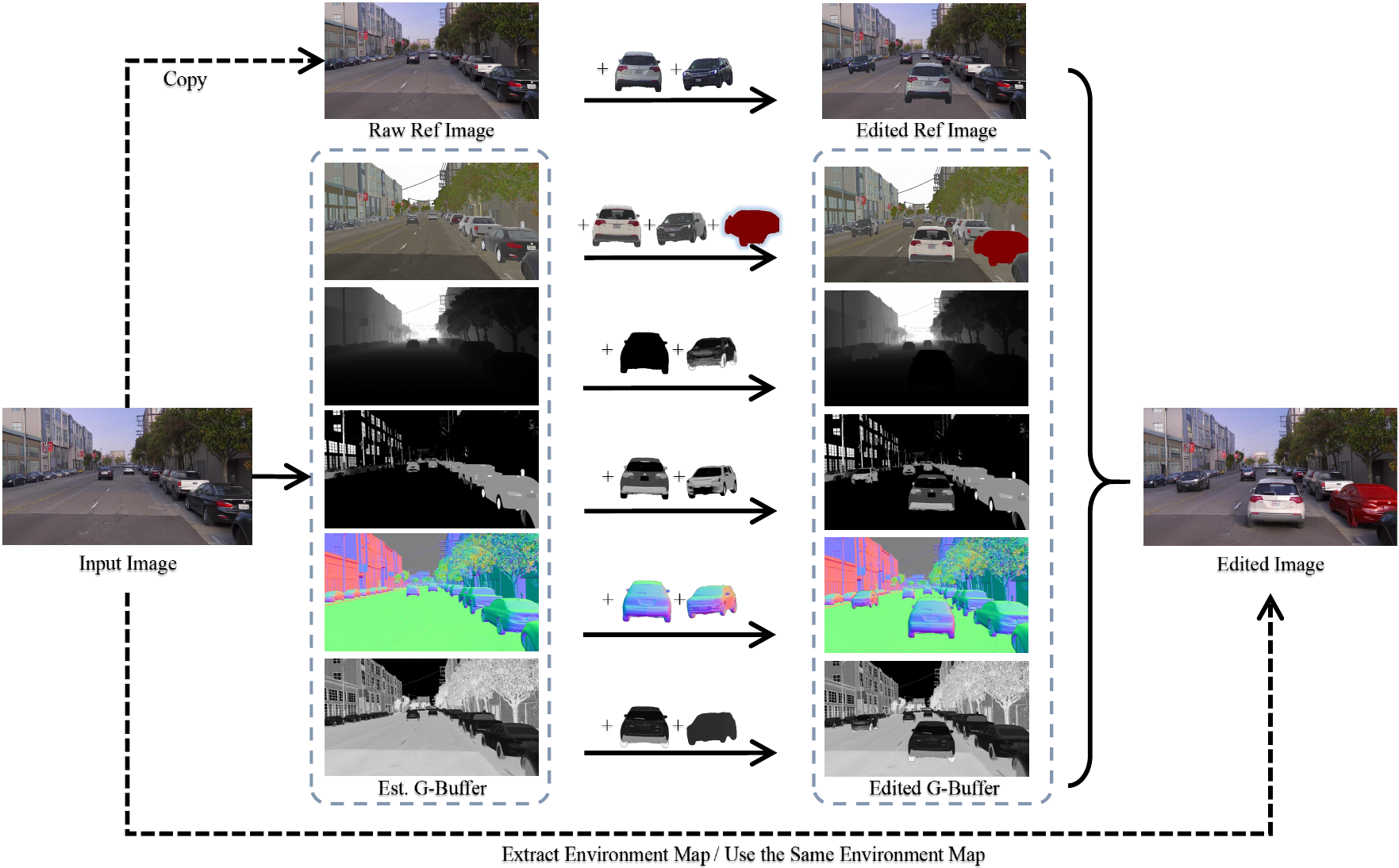}
  \caption{\textbf{Overview of the scene editing workflow.} Includes object insertion and material modification.}
   \label{fig:editing_detail}
\end{figure*}

We illustrate our scene editing workflow in Figure~\ref{fig:editing_detail}.
Specifically, we modify materials within the scene and insert new objects by editing intermediate intrinsics.
We utilize Ground-SAM~\cite{ren2024grounded} to obtain object masks, enabling material adjustments for specific objects.
Simultaneously, we directly insert new objects into the original image to serve as the reference image for model input.
Since the processed reference image still contains elements awaiting material modification, we employ latent space interpolation from Equation.~\ref{interpolation} to generate the edited image.

\subsection{Analysis of failure cases}

\begin{figure}[t]
  \centering
  \includegraphics[width=0.45\textwidth]{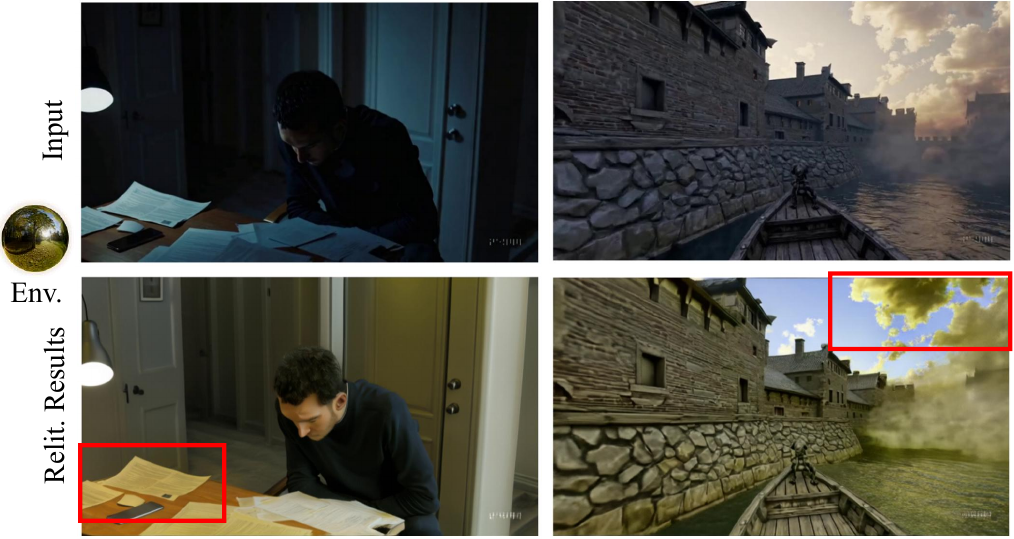}
  \caption{Failure cases of relighting: color shift (left) and abnormal illumination (right).}
   \label{fig:vis_failure}
\end{figure}

In this section, we focus on analyzing some typical failure cases of this method. As shown in Figure~\ref{fig:vis_failure}, some objects in the relighting results exhibit color shifts and abnormal lighting. We think these issues stem from inherent pseudo-label errors (G-buffer \& environment map) in the real-world training set. This is unavoidable for methods applied to real scenes, such as Diffusion Renderer. Inaccurate base-color labels may cause color shifts. Inaccurate environment map pseudo-labels impair the model's learning of light sources, causing it to occasionally misclassify background pixels as strong light sources.

\subsection{More visualizations of our methods}

\begin{figure*}[t]
\centering
\includegraphics[width=\textwidth]{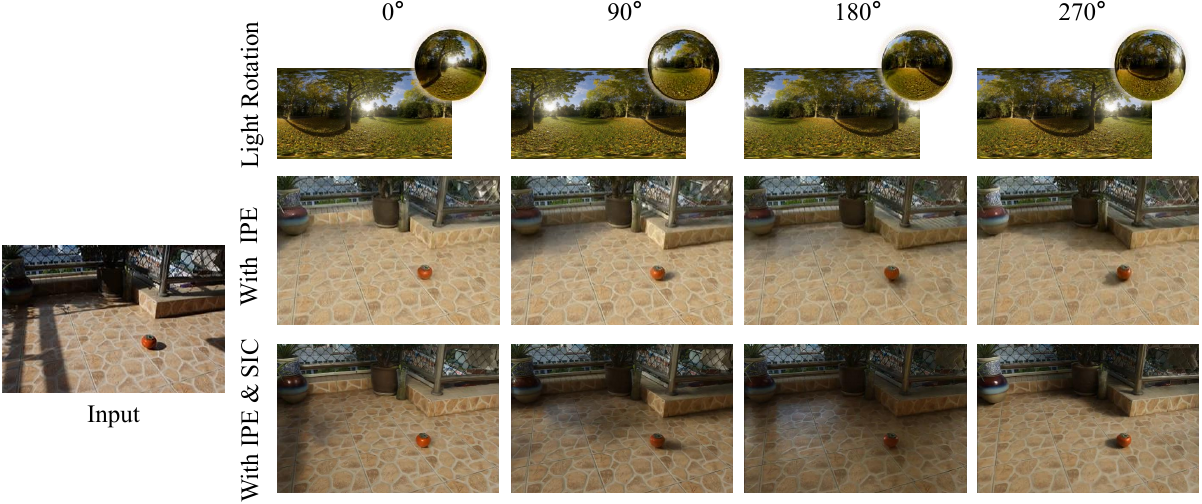}
\caption{\textbf{Ablation on training strategies.} IPE: Intrinsic Perception Enhancement. SIC: Self-supervised learning based on Illumination Consistency.}
\label{fig:append_training}
\end{figure*}

\begin{figure*}[t]
  \centering
  \includegraphics[width=\textwidth]{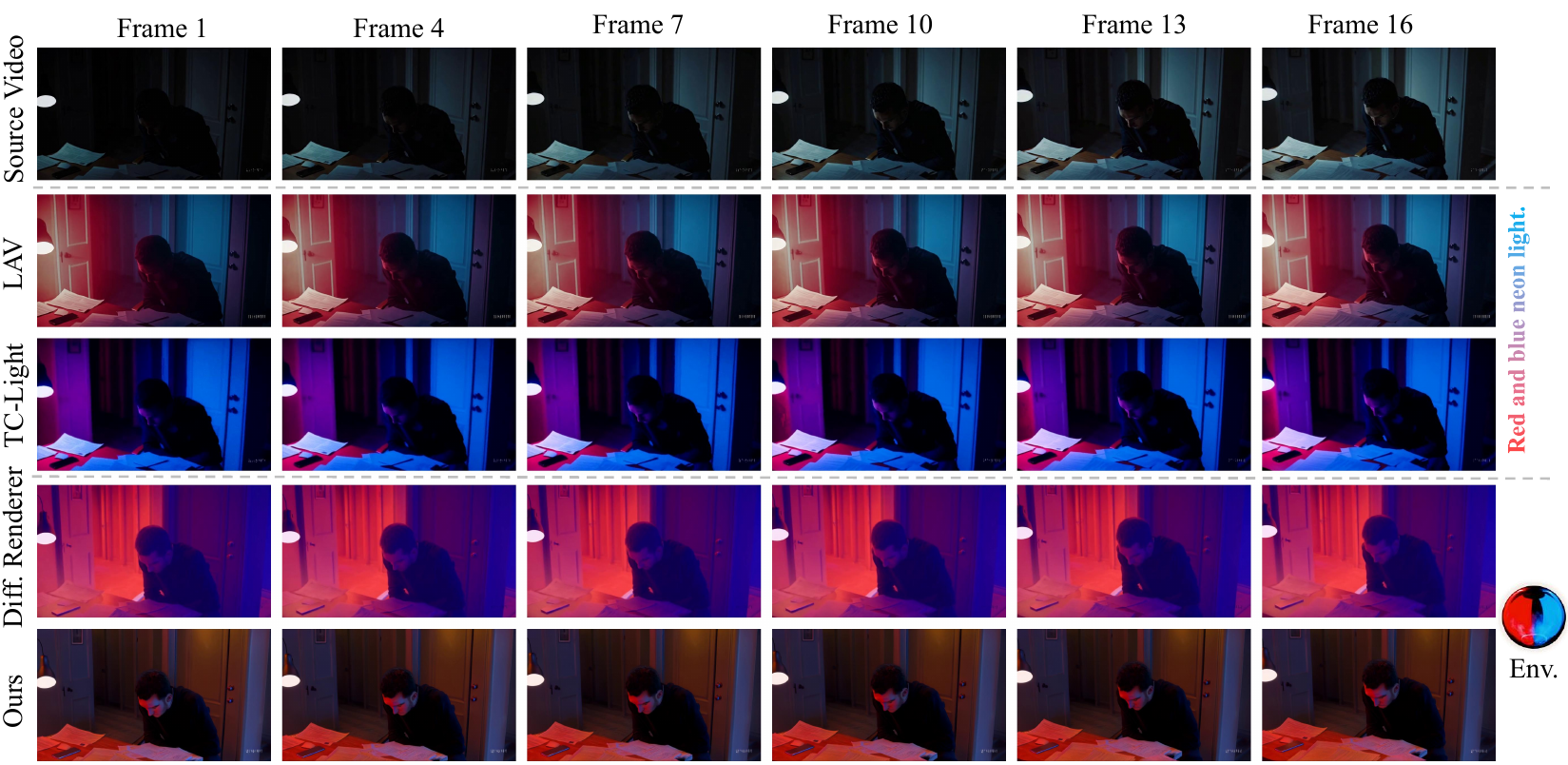}
  \caption{\textbf{Qualitative comparison of video relighting}. Our method achieves superior relighting quality, temporal consistency, and photorealistic generation results compared to baseline methods.}
  \label{fig:append_compare_1}
\end{figure*}

\begin{figure*}[t]
  \centering
  \includegraphics[width=\textwidth]{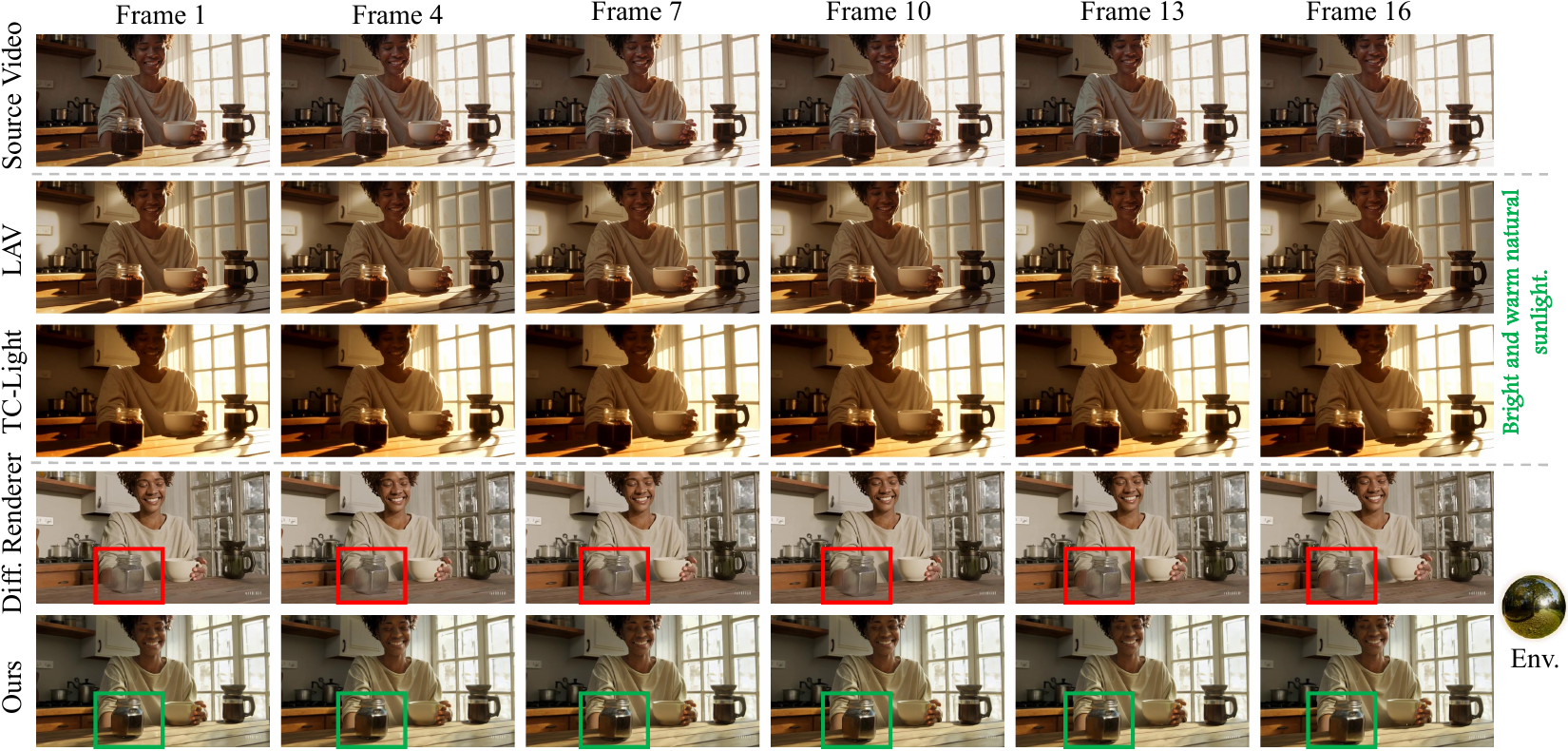}
  \caption{\textbf{Qualitative comparison of video relighting}. Our method achieves superior relighting quality, temporal consistency, and photorealistic generation results compared to baseline methods.}
  \label{fig:append_compare_2}
\end{figure*}

\begin{figure*}[t]
  \centering
  \includegraphics[width=\textwidth]{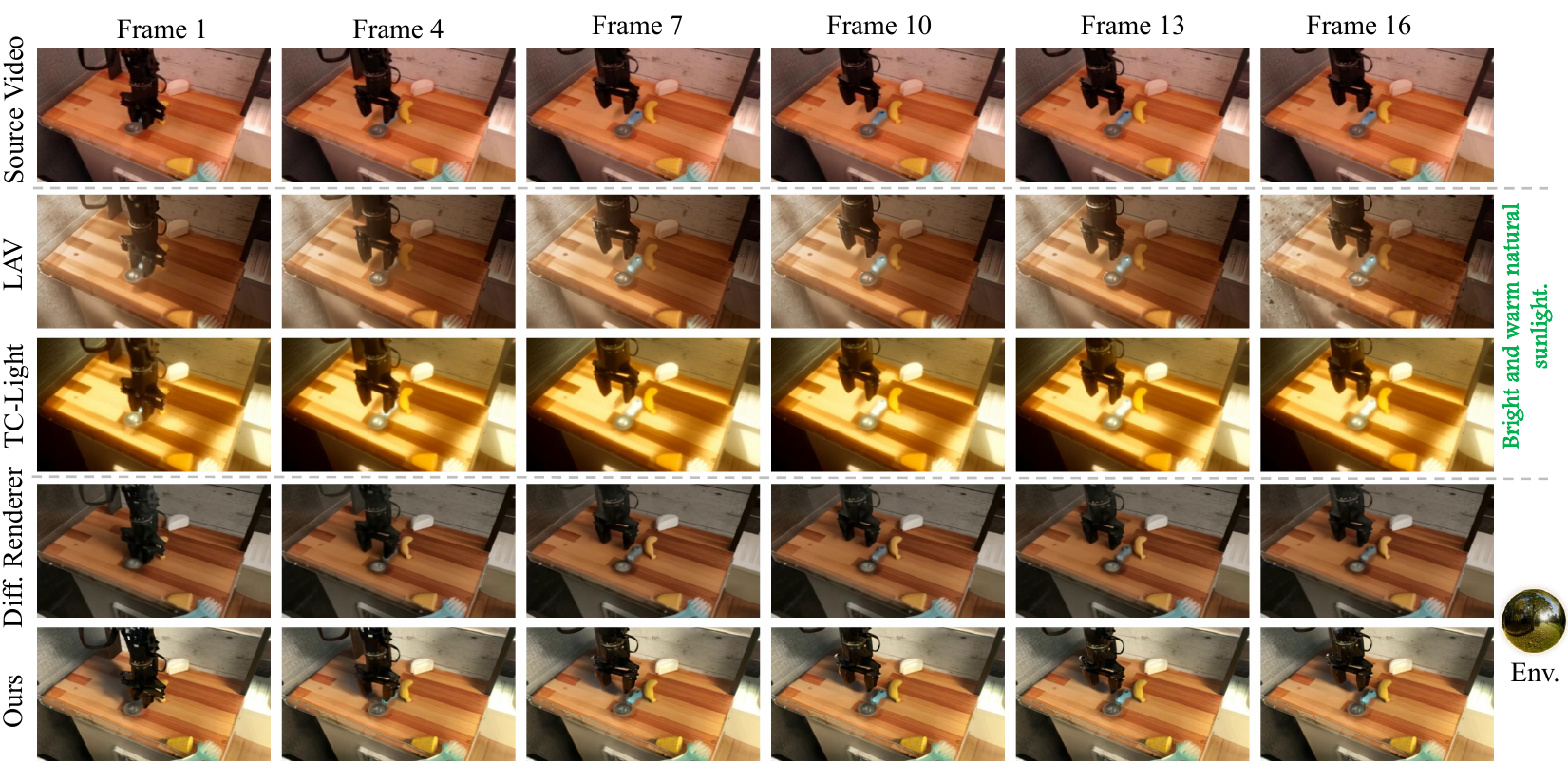}
  \caption{\textbf{Qualitative comparison of video relighting}. Our method achieves superior relighting quality, temporal consistency, and photorealistic generation results compared to baseline methods.}
  \label{fig:append_compare_3}
\end{figure*}

\begin{figure*}[t]
  \centering
  \includegraphics[width=\textwidth]{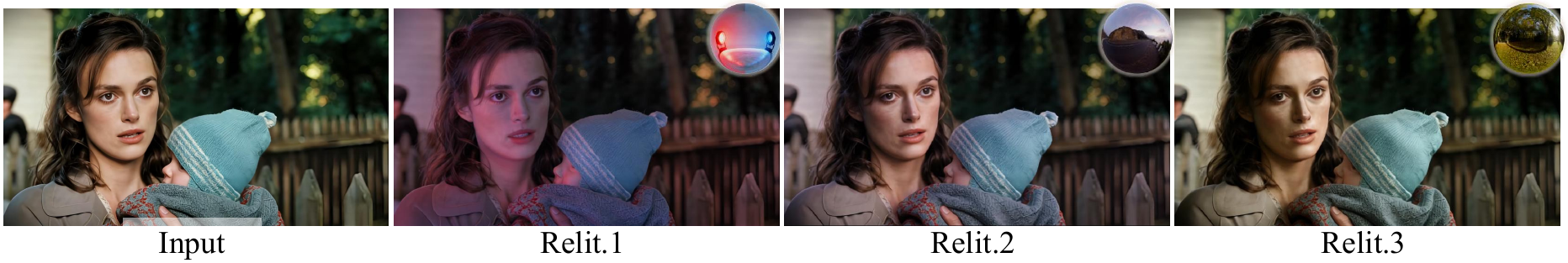}
  \caption{\textbf{Image relighting results of our method on portraits.}}
  \label{fig:append_ours_1}
\end{figure*}

\begin{figure*}[t]
  \centering
  \includegraphics[width=\textwidth]{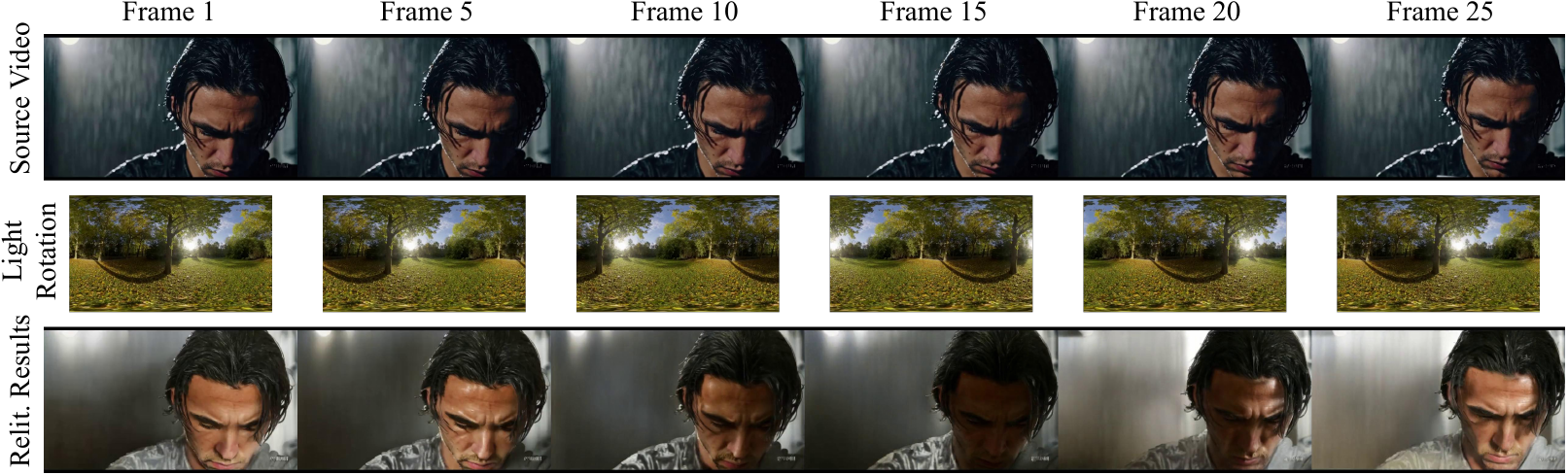}
  \caption{\textbf{Video results under dynamic lighting in a dynamic scene.}}
  \label{fig:append_ours_2}
\end{figure*}

\begin{figure*}[t]
  \centering
  \includegraphics[width=\textwidth]{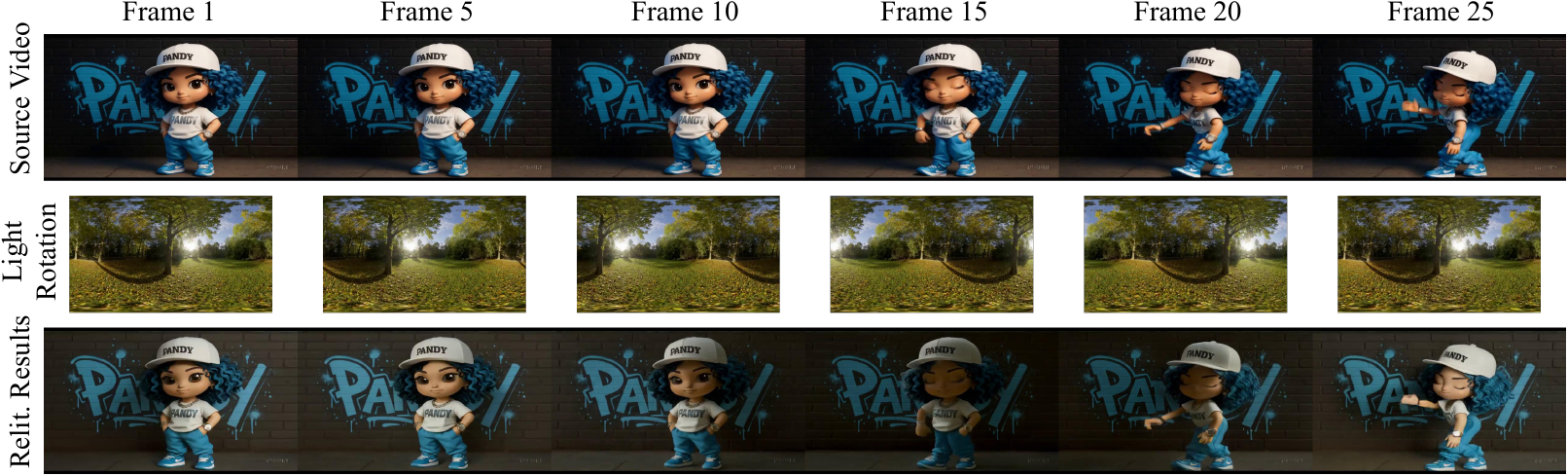}
  \caption{\textbf{Video results under dynamic lighting in a dynamic scene.}}
  \label{fig:append_ours_3}
\end{figure*}

\begin{figure*}[t]
  \centering
  \includegraphics[width=\textwidth]{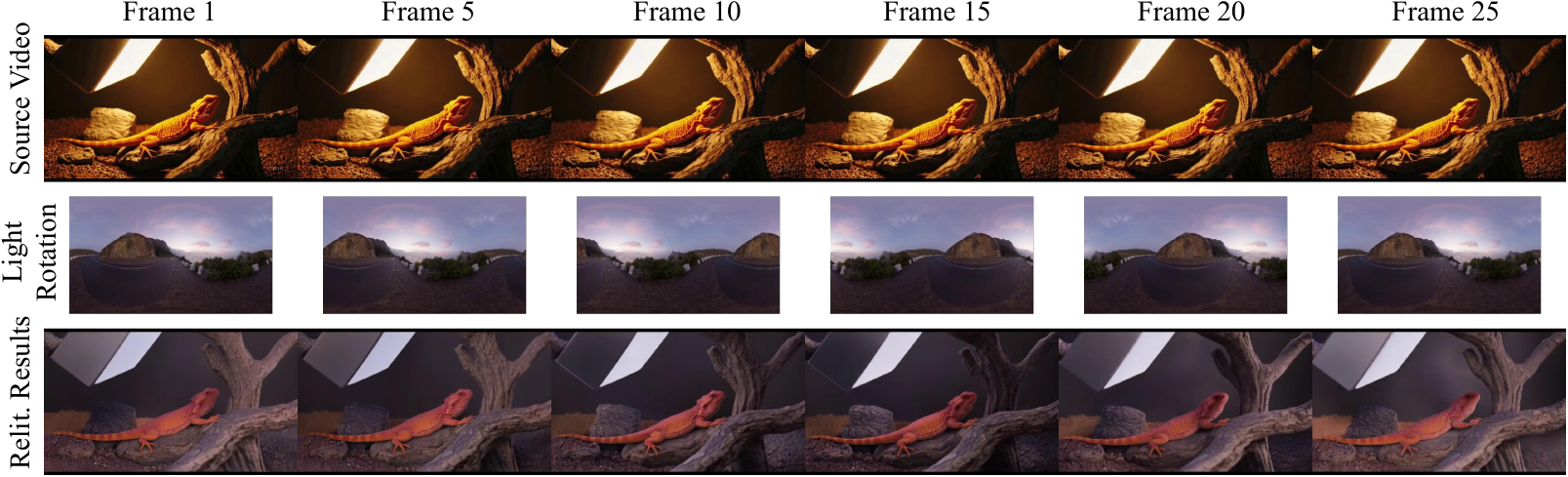}
  \caption{\textbf{Video results under dynamic lighting in a dynamic scene.}}
  \label{fig:append_ours_4}
\end{figure*}

\begin{figure*}[t]
  \centering
  \includegraphics[width=\textwidth]{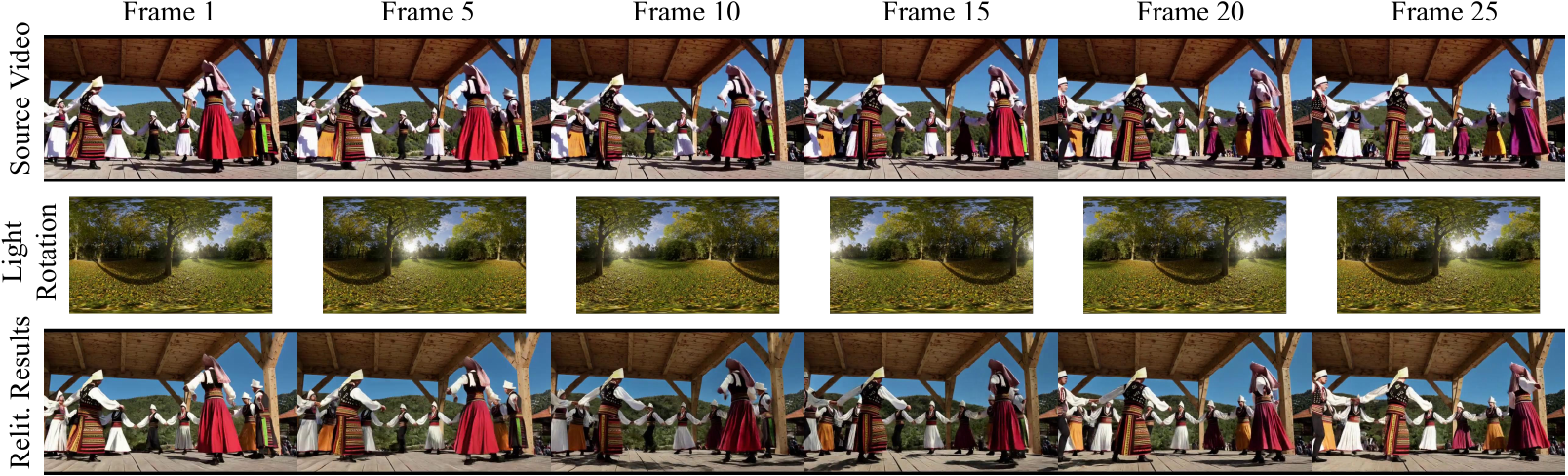}
  \caption{\textbf{Video results under dynamic lighting in a dynamic scene.}}
  \label{fig:append_ours_5}
\end{figure*}

\begin{figure*}[t]
  \centering
  \includegraphics[width=\textwidth]{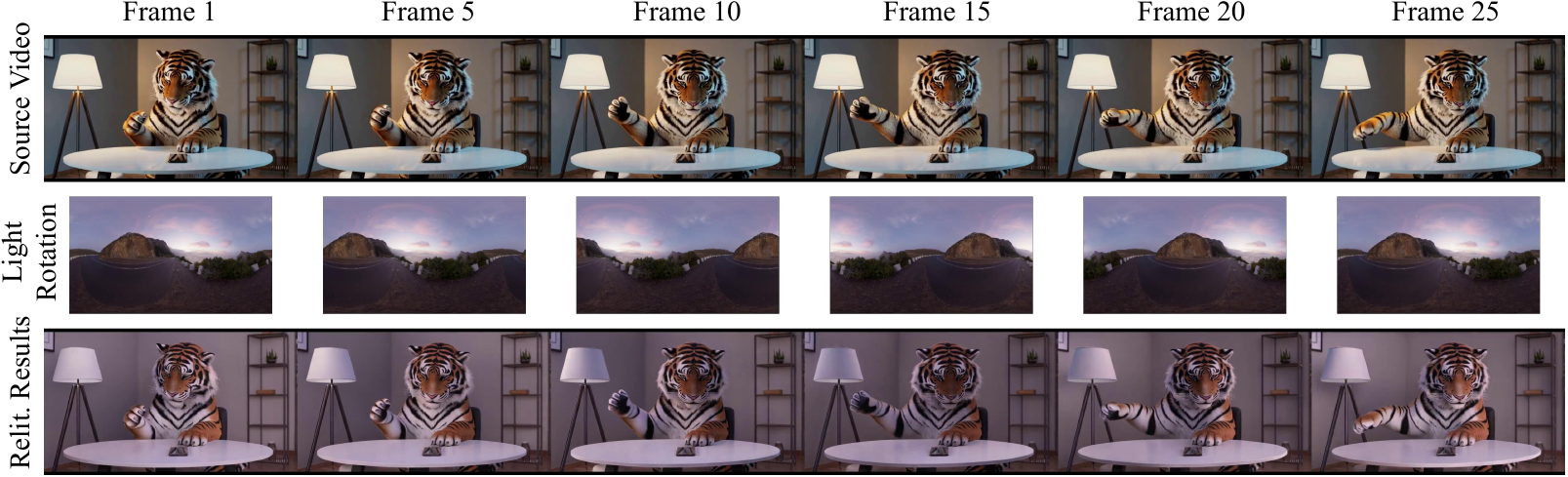}
  \caption{\textbf{Video results under dynamic lighting in a dynamic scene.}}
  \label{fig:append_ours_6}
\end{figure*}

\begin{figure*}[t]
  \centering
  \includegraphics[width=\textwidth]{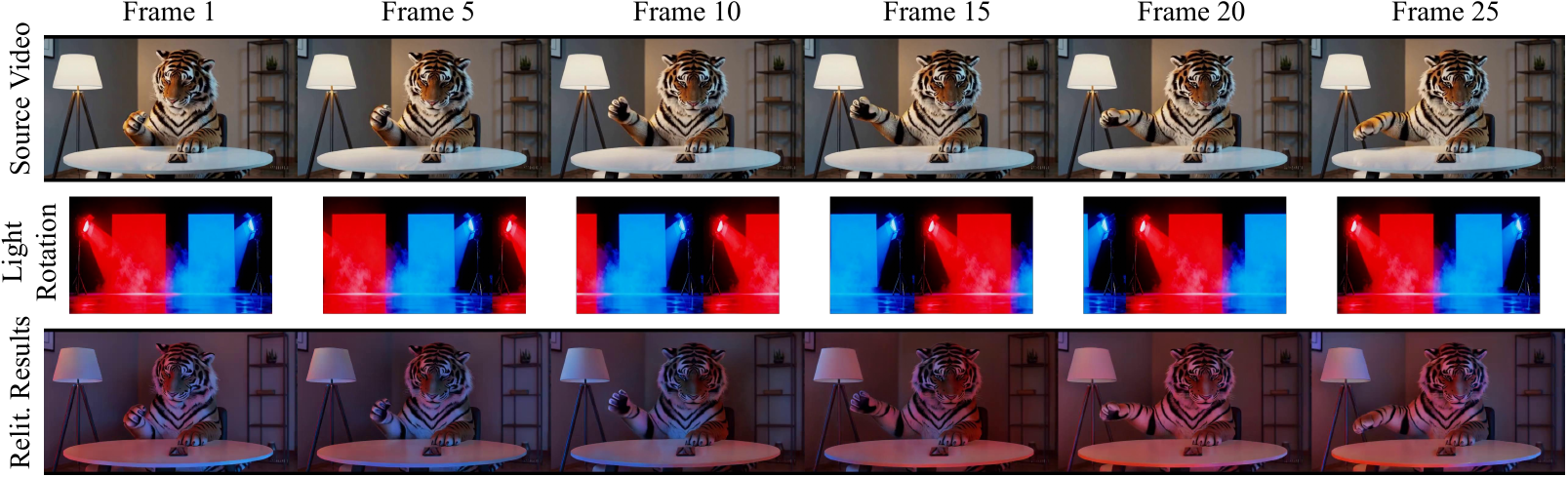}
  \caption{\textbf{Video results under dynamic lighting in a dynamic scene.}}
  \label{fig:append_ours_7}
\end{figure*}

\begin{figure*}[t]
  \centering
  \includegraphics[width=\textwidth]{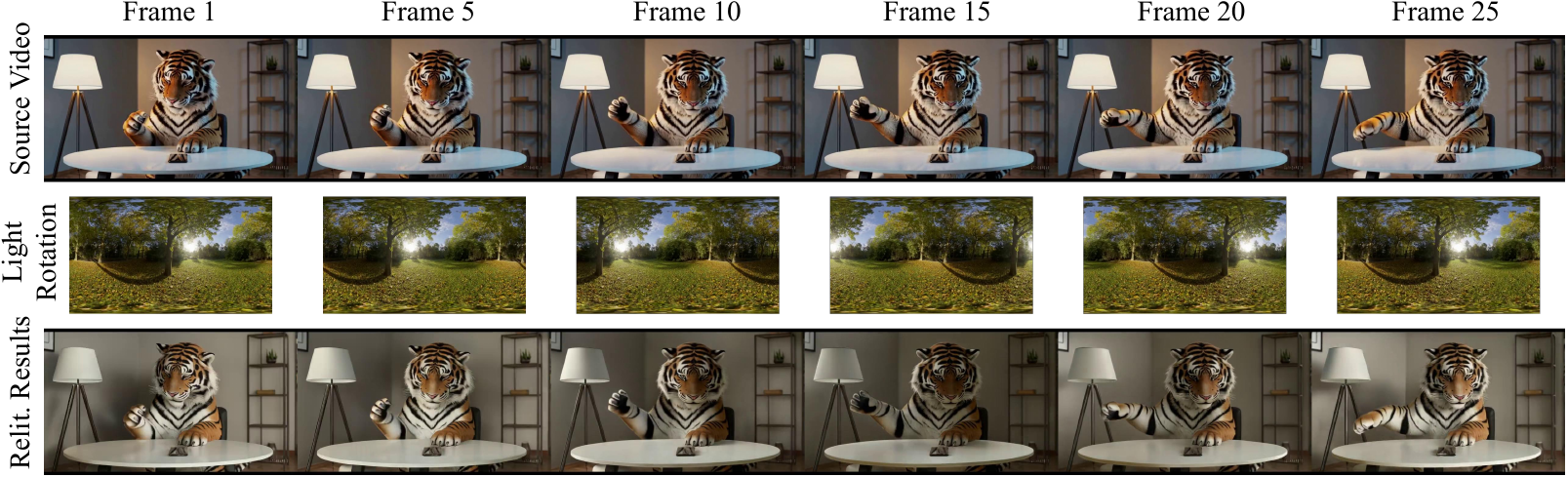}
  \caption{\textbf{Video results under dynamic lighting in a dynamic scene.}}
  \label{fig:append_ours_8}
\end{figure*}

\begin{figure*}[t]
  \centering
  \includegraphics[width=\textwidth]{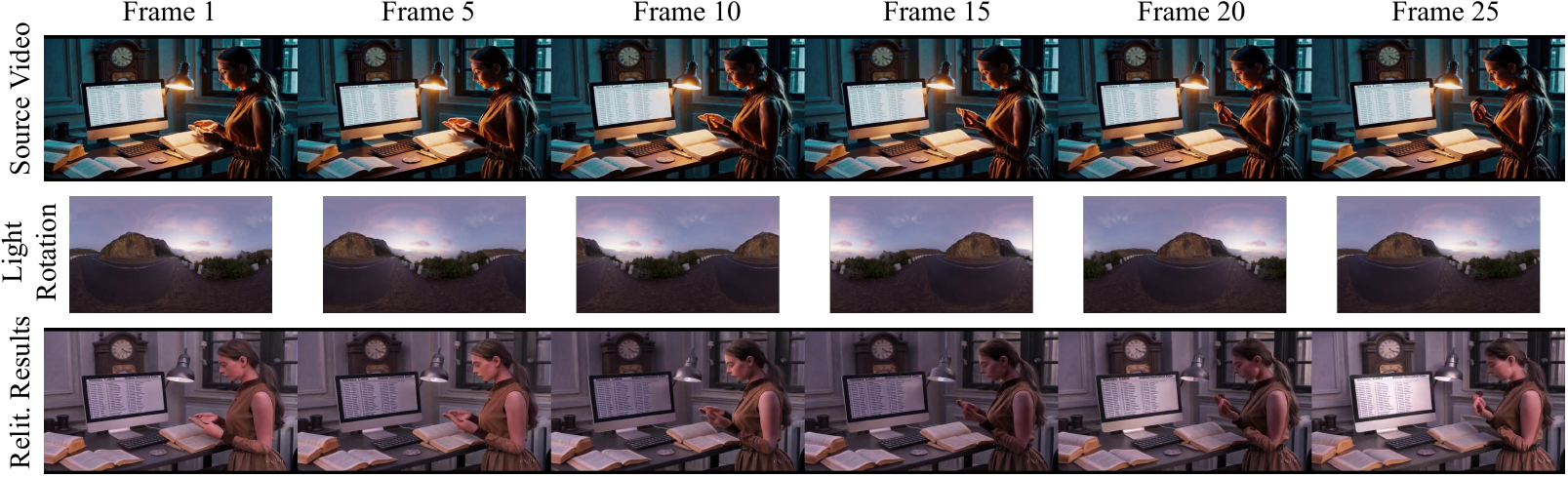}
  \caption{\textbf{Video results under dynamic lighting in a dynamic scene.}}
  \label{fig:append_ours_9}
\end{figure*}

\begin{figure*}[t]
  \centering
  \includegraphics[width=\textwidth]{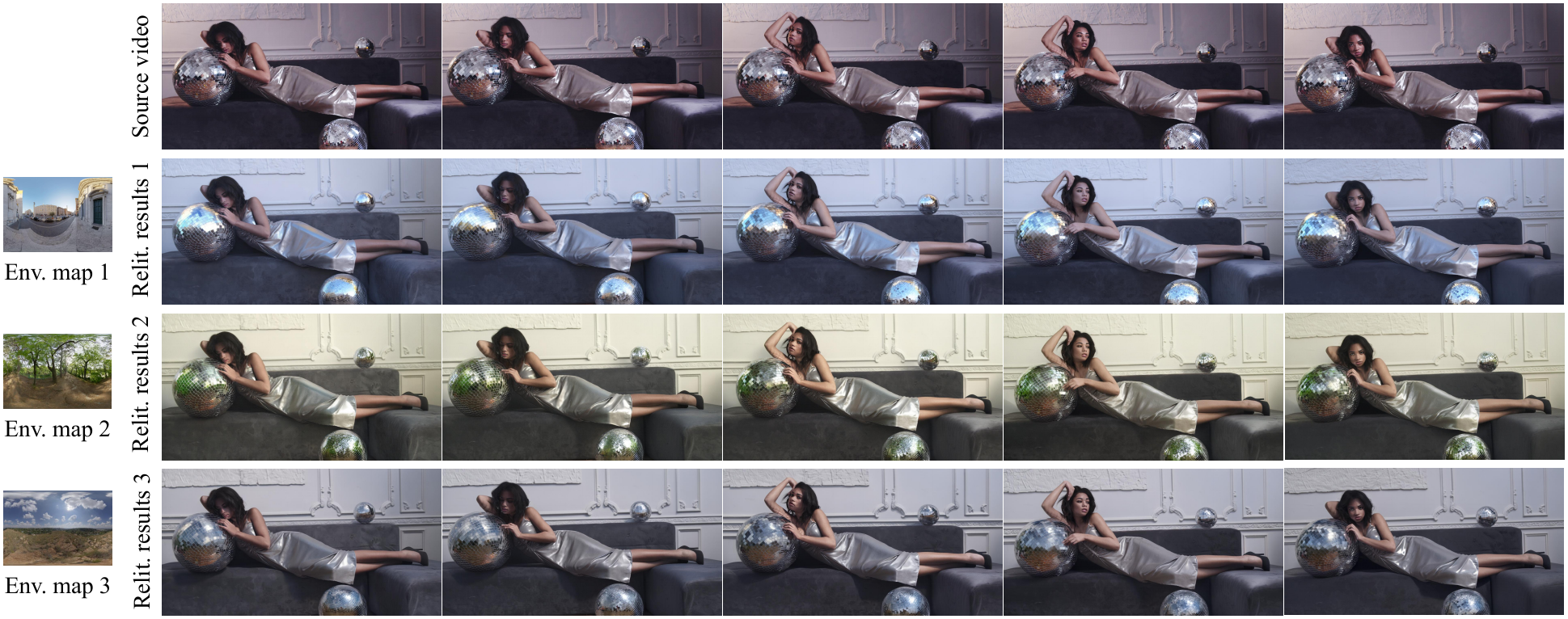}
  \caption{\textbf{Video results of the same scene under different environment lighting conditions.}}
  \label{fig:append_ours_10}
\end{figure*}

\begin{figure*}[h]
  \centering
  \includegraphics[width=\textwidth]{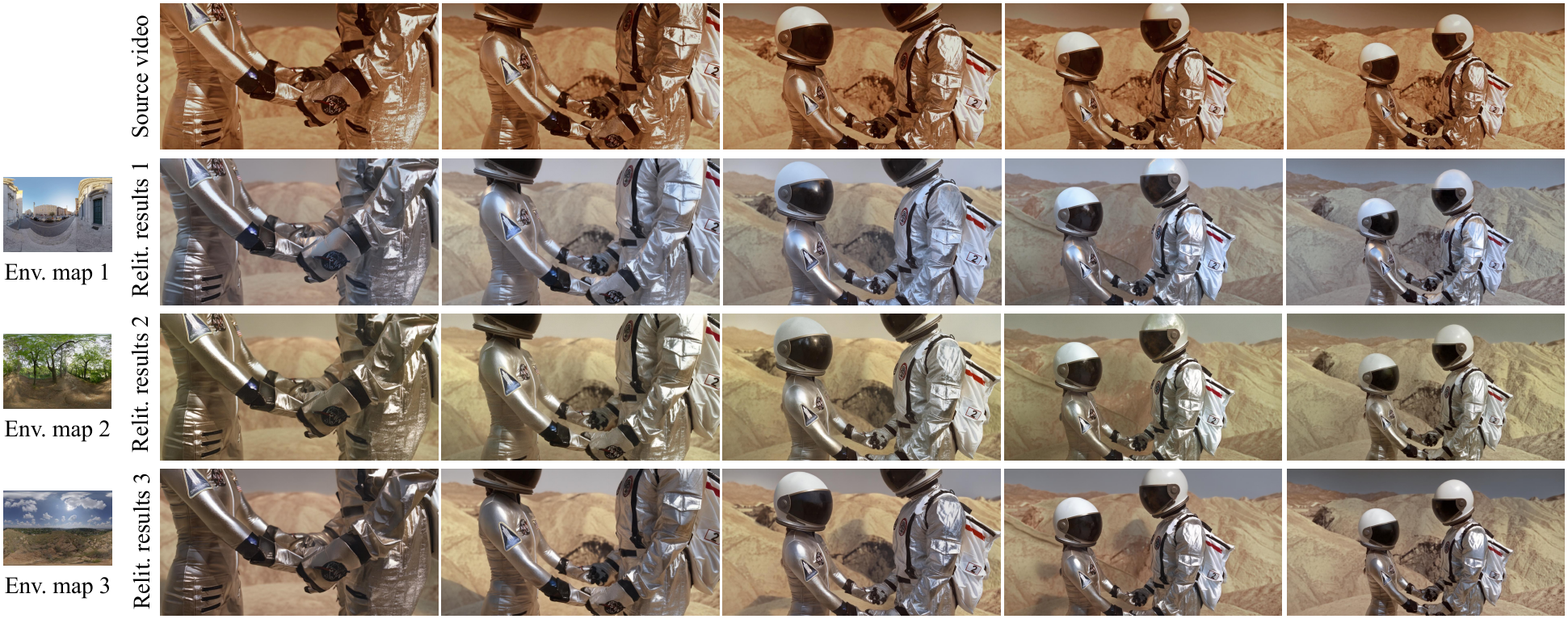}
  \caption{\textbf{Video results of the same scene under different environment lighting conditions.}}
  \label{fig:append_ours_11}
\end{figure*}

\begin{figure*}[h]
  \centering
  \includegraphics[width=\textwidth]{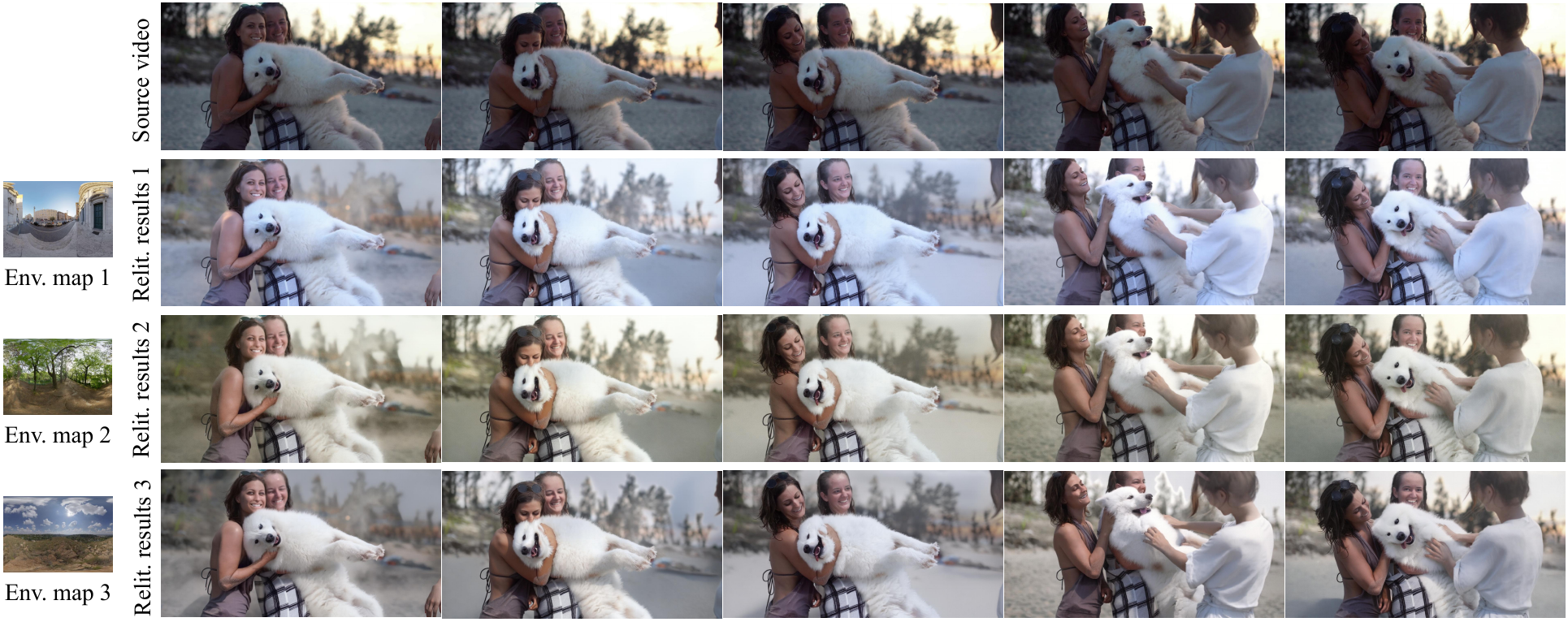}
  \caption{\textbf{Video results of the same scene under different environment lighting conditions.}}
  \label{fig:append_ours_12}
\end{figure*}

Figure~\ref{fig:append_training} visualizes the performance of our training strategy. Figures~\ref{fig:append_compare_1}-\ref{fig:append_compare_3} present additional visual comparisons of our method against other methods on in-the-wild data. Figures~\ref{fig:append_ours_1}-\ref{fig:append_ours_12} show the results of our method under dynamic lighting and various environment lights.

\end{document}